\documentclass[10pt,journal,twocolumn]{IEEEtran}

\ifCLASSINFOpdf
   \usepackage[pdftex]{graphicx}
\else
\fi

\usepackage{bm}
\usepackage{cite}
\usepackage{ifthen}
\usepackage{tikz-cd}
\usepackage{tikz}
\usepackage{amsmath}
\usepackage{amsfonts}
\usepackage{caption}
\usepackage{colortbl}
\usepackage{booktabs}
\usepackage{subfig}
\DeclareCaptionLabelSeparator{mysep}{\\ \vspace{0.15cm}}
\tikzset{every node/.append style = {draw=black,thin}}
\usetikzlibrary{trees}
\usetikzlibrary{shadings}
 \usetikzlibrary{backgrounds,scopes} 
\usepackage{xcolor}
\usepackage{multirow}
\usepackage{pstricks}
\usepackage{pst-sigsys}

\usepackage{xpatch}

\newboolean{draft}
\newcommand{\isdraft}[2]{\ifthenelse{\boolean{draft}}{#1}{#2}}

\isdraft{\renewcommand{\baselinestretch}{1.5}}{}

\definecolor{darkblue}{RGB}{0,0,100}	
\definecolor{darkredc}{RGB}{90,0,0}
\definecolor{thisblue}{rgb}{0.03, 0.27, 0.49}
\definecolor{darkgreen}{rgb}{0.0, 0.42, 0.24}

\newcommand{\mypar}[1]{{\bf #1.}}

\begin{document}


\title{Robust Single-Image Super-Resolution via CNNs and TV-TV Minimization}

\author{Marija Vella \,%
         and\,\,
         Jo\~{a}o F. C. Mota
\thanks{Part of this work has been presented in \cite{vella_single_2019,vella_TV_2019}.
  Marija Vella and Jo\~{a}o F. C. Mota are with the School of Engineering \& Physical Sciences, Heriot-Watt University, Edinburgh EH14 4AS,
UK. (e-mail: \{mv37, j.mota\}@hw.ac.uk).}
\thanks{}}


\maketitle

\begin{abstract}

Single-image super-resolution is the process of increasing the resolution of an image, obtaining a high-resolution (HR) image from a low-resolution (LR) one. By leveraging large training datasets, convolutional neural networks (CNNs) currently achieve the state-of-the-art performance in this task. Yet, during testing/deployment, they fail to enforce consistency between the HR and LR images: if we downsample the output HR image, it never matches its LR input. Based on this observation, we propose to post-process the CNN outputs with an optimization problem that we call \textit{TV-TV minimization}, which enforces consistency. As our extensive experiments show, such post-processing not only improves the quality of the images, in terms of PSNR and SSIM, but also makes the super-resolution task robust to operator mismatch, i.e., when the true downsampling operator is different from the one used to create the training dataset.   


\end{abstract}


\isdraft{\newpage}{}

%
\IEEEpeerreviewmaketitle

\section{Introduction}
\label{Section1}

\IEEEPARstart{I}{n} science and engineering, images acquired by
sensing devices often have resolution well below the desired one. Common reasons
include physical constraints, as in astronomy or biological microscopy, and
cost, as in consumer electronics or medical imaging. Creating high-resolution
(HR) images from low-resolution (LR) ones, a task known as
\textit{super-resolution} (SR), can therefore be extremely useful in these
areas; it enables, for example, the identification of structures or objects
that are barely visible in LR images. Doing so, however, requires inferring
values for the unobserved pixels, which cannot be done without making
assumptions about the class of images to super-resolve and their
acquisition process. 

Classical interpolation algorithms assume that the missing pixels can be
inferred by linearly combining neighboring pixels via, e.g., the application of
filters~\cite{blu_linear_2004,keys_cubic_1981}, or by preserving the statistics
of the image gradient from the LR to the HR image~\cite{fattal_image_2007}.
Reconstruction-based methods, on the other hand, assume that images have sparse
representations in some domain, e.g., sparse
gradients~\cite{Osher_1990,rudin_nonlinear_1992,li_efficient_2013,morse_image_2001,aly_image_2005,chambolle_algorithm_2004,bioucas-dias_new_2007,becker_nesta:_2011}.
More recently, data-driven methods have become very popular; their main
assumption is that image features can be learned from training data, via
dictionaries~\cite{yang_image_2010,singh_super-resolving_2014} or via
convolutional neural networks
(CNNs)~\cite{dong_learning_2014,wang2018esrgan,Kim_VDSR}.

CNNs were first applied to image SR in the seminal
work~\cite{dong_learning_2014} and have ever since remained the
state-of-the-art, both in terms of reconstruction performance and computational
complexity (during deployment/testing). By relying on vast databases of images
for training, such as ImageNet \cite{Deng09imagenet} or T91
\cite{yang_image_2010}, they can effectively learn to map LR images/patches to
HR images/patches. Although training a CNN can take several days, applying it
to an image (what is typically called the testing phase) takes a few seconds
or even sub-seconds. Despite these advantages, the knowledge that CNNs extract
from data is never made explicit, making them hard to adapt to new scenarios:
for example, simply changing the scaling factor or the
sampling model, e.g., from bicubic to point sampling, almost always requires
retraining the entire network. More conspicuously, however, is that during
testing SR CNNs fail to guarantee the consistency between the reconstructed HR
image and the input LR image, effectively ignoring precious ``measurement"
information, as we will illustrate shortly. Ignoring such information makes
CNNs prone to generalization errors and, as a consequence, also less robust.

Curiously, adaptability and measurement consistency are the main features of
classical reconstruction-based methods, which consist of algorithms designed
to solve an optimization problem. To formulate such an optimization problem,
one has to explicitly encode the measurement model and the assumptions
about the class of images to be super-resolved. Although this explicit encoding
confers reconstruction-based methods great adaptability and flexibility, it
naturally limits the complexity of the assumptions, which is one of
the reasons why reconstruction-based methods are outperformed by data-driven
methods (CNNs). This motivates our problem: 

\mypar{Problem statement}
\textit{Can we design SR algorithms that learn from large quantities of data
and, at the same time, are easily adaptable to new scenarios and guarantee
measurement consistency during the testing phase?} In other words, can we
design algorithms that have the advantages of both data-driven and model-based
methods?

\begin{figure}[!t]
\centering
\includegraphics[width=9cm,height = 4.8cm]{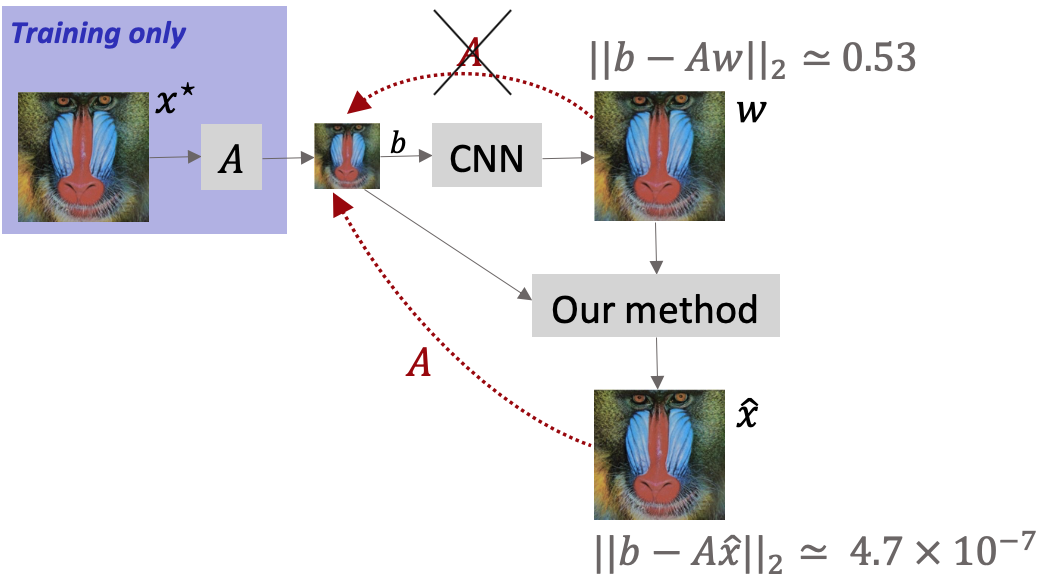}
\caption{
Illustration of the lack of measurement consistency by CNNs during
\textit{testing}: when the output image $w$ is downsampled using $A$, it
typically differs significantly from $b$. Our method takes in both $w$ and $b$, and
fixes this problem.}
\label{fig_prob}
\end{figure}

\mypar{Lack of consistency by CNNs} Before summarizing our method, we describe
how CNNs fail to enforce consistency between the reconstructed HR image and the
input LR image. Although we illustrate this phenomenon here for the specific
SRCNN network~\cite{dong_learning_2014}, more systematic experiments can be
found in Section \ref{Section4}. The top-left corner of Fig.\ \ref{fig_prob}
shows a ground truth (GT) image $X^\star \in \mathbb{R}^{M \times N}$, which
the algorithms have access to during training, but not during testing.\footnote{Although the figure displays
  color images, most SR algorithms work only on the luminance channel of a
YCbCr representation (or a grayscale image).  Color images can be obtained a
posteriori by adding the remaining channels.} We will represent $X^\star$
by its column-major vectorization $x^{\star} = \text{vec} (X^{\star}) \in
\mathbb{R}^n$, where $n = M\cdot N$. The GT image $x^{\star}$ is downsampled
via a linear operator represented by $A\in\mathbb{R}^{m \times n}$ into a LR
image  $b := A{x^{\star}}\in\mathbb{R}^m$. In this specific example,
$n=240,000$ and $m=15,000$, i.e., $x^\star$ is downsampled by factor of 4,
and $A$ implements bicubic downsampling. The goal is to reconstruct $x^{\star}$
from $b$, i.e., to super-resolve $b$. The figure illustrates that CNN methods,
in particular \cite{dong_learning_2014}, reconstruct an image $w$ that does not
necessarily satisfy $Aw = b$, even though we know that $Ax^{\star} = b$.
Specifically, SRCNN \cite{dong_learning_2014} outputs an image $w = \text{vec}
(W)$ (top-right corner) very close to $x^{\star}$ (22.73 dBs in PSNR) but that
fails to satisfy $Aw=b$ with enough precision: $||b - Aw||_2 \simeq
0.53$. We point out that $A$ represents bicubic
downsampling, which was what the authors of \cite{dong_learning_2014} assumed
during the training of SRCNN.

\mypar{Our approach} 
Our algorithm can be viewed as a post-processing step that takes as input
the CNN image $w$ and the LR image $b$, and reconstructs a HR image $\hat{x}
\in \mathbb{R}^n$ that is not very dissimilar from $w$ but, in contrast to it,
satisfies $A\hat{x} = b$ (bottom-right corner of Fig. \ref{fig_prob}). As a
result, the images created by our method almost always have better quality than
$w$, in terms of PSNR and SSIM. In addition, our method confers robustness to
the SR task, even in the case where the operator $A$ used to generate the
training data differs from the one used during testing.
 
We integrate $b$ and $w$ via an algorithm that solves an optimization problem
that we call \textit{TV-TV minimization}. The problem enforces the
reconstructed image to have a small number of edges, a property captured by a
small TV-norm, and also to not differ much from $w$, as measured again by the
TV-norm. Naturally, it also imposes the constraint $A\hat{x} = b$. 

\mypar{Contributions} 
We summarise our contributions as follows:
\begin{enumerate}
  \item We introduce a framework that has the advantages of learning-based
    and reconstruction-based methods. Like reconstruction methods, it
    is adaptable, flexible, and enforces measurement consistency. At the same time, it
    retains the excellent performance of learning methods.

  \item We integrate learning and reconstruction-based methods via a TV-TV
    minimization problem. Although we have no specific theoretical guarantees
    for it, existing theory for a related, simpler problem
    ($\ell_{1}$-$\ell_{1}$ minimization) provides useful insights about how to
    tune a regularization parameter. This makes our algorithm easy to deploy,
    since there are virtually no parameters to tune.

  \item We propose an algorithm based on the alternating direction method of
    multipliers (ADMM) \cite{boyd_distributed_2011} to solve the TV-TV
    minimization problem. In contrast with most SR methods, which process image
    patches independently, our algorithm processes full images at once. It also
    easily adapts to different degradations and scaling factors.

  \item We conduct extensive experiments that illustrate not only the
    robustness of our algorithm under different degradation operators, but
    also how it systematically improves (in terms of PSNR and SSIM) the outputs
    of state-of-the-art SR networks, such as ESRGAN \cite{wang2018esrgan}, VDSR \cite{Kim_VDSR} and LapSRN \cite{lai_deep_2017}.
\end{enumerate}

We highlight the following differences with respect to our previous work in
\cite{vella_single_2019,vella_TV_2019}. We now explore and illustrate with
experiments the underlying reason why our framework improves the output of
state-of-the-art SR CNNs. We also describe how the proposed optimization
problem can be solved efficiently using ADMM; in fact, the algorithms used in
\cite{vella_single_2019,vella_TV_2019} were different and less efficient than
the algorithm we present here. Our experiments are also much more extensive:
they consider different sampling operators to illustrate robustness to operator mismatch, and include many more algorithms, e.g., FSRCNN
\cite{dong_accelerating_2016}, VDSR \cite{Kim_VDSR}, LapSRN
\cite{lai_deep_2017}, SRMD \cite{zhang_learning_2018}, IRCNN
\cite{IRCNN-zhang2017}, and ESRGAN \cite{wang2018esrgan}.

\mypar{Organization}
Section~\ref{Section2} summarizes prior work on SR algorithms, and
Section~\ref{Section3} describes the proposed framework and optimization
scheme. Section \ref{Section4} then reports our experimental 
results, and Section \ref{Section5} concludes the paper. 

\section{Related Work} 
\label{Section2}

Super-resolution (SR) schemes are often labeled as interpolation,
reconstruction, or data-driven. Interpolation methods infer the missing pixels
by locally applying an interpolation function such as the bicubic or bilinear
filter~\cite{blu_linear_2004,keys_cubic_1981}. As they have been surpassed by
both reconstruction and learning SR algorithms, we will limit our review to the
latter.

 
\subsection{Reconstruction-Based SR}

Reconstruction-based schemes view SR as an image reconstruction problem and
address it by formulating an optimization problem. In general, the optimization
problem has two terms: a data consistency term that encodes assumptions about
the acquisition process, usually that $Ax \simeq b$ (where $x$ is the
optimization variable), and a regularization term on $x$ that encodes assumptions
about the class of images. Different methods differ mostly on the image
assumptions.

\mypar{Image assumptions} 
Reconstruction SR methods encode assumptions about the images by penalizing in
the optimization problem measures of complexity. These reflect the empirical
observation that natural images have parsimonious representations in several
domains.  Examples include sparsity in the wavelet
domain~\cite{wavelet_Mallat_2010}, sparsity of image patches in the DCT
domain~\cite{DCT_hallucination_2011} and, as we will explore shortly in more
detail, sparsity of image
gradients~\cite{Osher_1990,rudin_nonlinear_1992,li_efficient_2013,morse_image_2001,aly_image_2005,chambolle_algorithm_2004,bioucas-dias_new_2007,becker_nesta:_2011}.
Since sparsity is well captured by the $\ell_1$-norm, the resulting
optimization problem is typically convex and can be solved efficiently. A more
challenging assumption is multi-scale
recurrence~\cite{Glasner_2009,Lim_2017_CVPR_Workshop}, which captures the
notion that patches of natural images occur repeatedly across the image. For
example, \cite{Glasner_2009} proposed an SR algorithm that explores the
recurrence of image patches both in the same and in different scales.



\mypar{Total variation} 
In natural images, the number of pixels that correspond to an edge, i.e.,
a transition between different objects, is a small percentage of the total
number of pixels. This can be measured by the total variation (TV) of
the image~\cite{Osher_1990}. Although TV was initially defined in the context of partial
differential equations, there has been work that discretizes the differential
equations~\cite{morse_image_2001,aly_image_2005} or that directly defines TV in
the discrete
setting~\cite{chambolle_algorithm_2004,bioucas-dias_new_2007,Condat17-DiscreteTotalVariation-NewDefinitionAndMinimization}.
Although there are several definitions of discrete TV, the most popular are the
\textit{isotropic TV}, which consists of the sum (over all pixels) of the
$\ell_2$-norms of the vectors containing the horizontal and vertical
differences at each pixel, and the \textit{anisotropic TV}, which is similar to
isotropic TV but with the $\ell_2$-norms replaced by the $\ell_1$-norm. Both
definitions yield convex, yet nondifferentiable, functions. Many algorithms
have been proposed to solve problems involving discrete TV, including
primal-dual methods~\cite{goldstein_split_2009,li_efficient_2013}, and proximal
and gradient-based
schemes~\cite{chambolle_algorithm_2004,bioucas-dias_new_2007,becker_nesta:_2011}.

The concept of TV has been used in many imaging tasks, from
denoising~\cite{Osher_1990,chambolle_algorithm_2004,Condat17-DiscreteTotalVariation-NewDefinitionAndMinimization}
to
super-resolution~\cite{morse_image_2001,chambolle_algorithm_2004,aly_image_2005}.
For example, \cite{morse_image_2001} discretizes a differential equation
relating variations in the values of pixels to the curvature of level sets,
while enforcing fidelity to the LR image. The work in \cite{aly_image_2005}
proposed a similar method, but with more complex models for both TV and the
image acquisition process.

\subsection{Learning-based Algorithms}
     
Learning-based algorithms typically consist of two stages: \textit{training}, in
which a map from LR to HR patches is learned from a database of training
images, and \textit{testing}, in which the learned map is applied to
super-resolve an unseen image. 

\mypar{Manifold learning}  
Manifold learning relies on the observation that most data (e.g., patches of
images) lie on a low-dimensional manifold. An example is locally linear
embedding~\cite{roweis_nonlinear_2000} which, like PCA, learns a compact
representation of data, but considers nonlinear geometry instead. This idea was
applied to SR in~\cite{chang_super-resolution_2004} by assuming that LR and HR
patches lie on manifolds with similar geometry. Thus, given a set of training
HR and LR patches and a target LR image, the method
in~\cite{chang_super-resolution_2004} first finds the LR patches in the
training set that are neighbors of the target LR patch. Then, it uses the HR
version of those neighbors to infer the HR target patches. The resulting
algorithm can over- or under-fit the data depending on how many nodes (i.e.,
patches) define a neighborhood.


       
\mypar{Dictionary learning} 
In dictionary learning, also known as sparse coding, patches of HR
images are assumed to have a sparse representation on an
over-complete dictionary, which is learned from training images. 
For example, \cite{yang_image_2010} uses training images to learn dictionaries
for LR and HR patches while constraining corresponding patches to have the same
coefficients. Other schemes use similar concepts, but require no training
data at all. For example, \cite{singh_super-resolving_2014} uses
self-similarity to learn the LR-HR map without any external database of images.
                  
\mypar{CNN-based methods} 
The advent of deep learning and the availability of large image datasets
inspired the application of CNNs to SR. Currently, they surpass any
reconstruction- or interpolation-based method both in reconstruction
performance and in execution time (during testing). The first CNN for SR was
proposed by \cite{dong_learning_2014}; although its design was inspired by
dictionary learning methods, the proposed architecture set a new standard for
SR performance.
        
SR networks can be classified as direct or progressive. In direct networks, the
LR image is first upscaled, typically via bicubic interpolation, to the
required spatial resolution, and then is fed to a CNN, as in
\cite{dong_learning_2014,dong_accelerating_2016, Tong_2017, Kim_VDSR}. In this
case, the CNN thus learns how to deblur the upscaled image. As previously
mentioned, CNN architectures need to be retrained every time we change the
scaling factor. To overcome this, \cite{kim_DRCN} repeatedly applied a
recursive convolutional layer to obtain the super-resolved image. However,
since the LR input is blurry, the CNN outputs a HR image lacking fine details.
Inspired by this observation, \cite{ledig_photo-realistic_2017} proposed the
SRGAN, which produces photo-realistic HR images, even though they do not yield
the best PSNR. As direct networks operate on high-dimensional images,
their training is computationally expensive~\cite{Shi2016RealTimeSI}. 
        
Progressive networks, in contrast, have reduced training complexity, as they
directly process LR images. Specifically, the upsampling step, performed using
sub-pixel or transposed convolution~\cite{Shi2016RealTimeSI}, is applied
only at the end of the network. For instance, LapSRN~\cite{lai_deep_2017}
used the concept of Laplacian pyramids, in which each network level is trained
to predict residuals between the upscaled images at different levels in the
pyramid. More recently, \cite{proSR} proposed a fully progressive network that
super-resolves images by an upsampling factor of two at each level until the
desired factor is reached. 
        
In spite of achieving state-of-the-art performance, CNNs for SR suffer from two
major shortcomings: as already illustrated, they fail to guarantee the consistency between the LR and
HR image during testing, and the trained network
applies only to a unique scaling factor and degradation function. Most of the
CNNs, e.g. \cite{dong_learning_2014, dong_accelerating_2016,
zhang_learning_2018}, are trained by solving
\begin{equation*}
  \underset{\theta}{\text{minimize}} \,\,\,
  \frac{1}{T} \sum_{t=1}^T \big\|f_{\theta}(Ax^{(t)}) - x^{(t)}\big\|_2^2\,,
\end{equation*}
where $x^{(t)}$ represents (the vectorization of) the $t$th image in the
training set, $A$ the bicubic sampling operator, and $f_\theta(\cdot)$ a CNN
parameterized by $\theta$ (i.e., weights and biases of the neural connections).
Most CNNs are trained with images that have been downsampled with a bicubic filter.
Thus, their performance quickly degrades when the true downsampling operator is
different. Indeed, during testing, the true degradation is unknown. 
The work in~\cite{zhang_learning_2018} addresses this
problem by designing a network that deals with different
degradations by accepting as input both the blur kernel and the noise level.

\subsection{Plug-and-Play Methods}
        
A different line or work blends learning- and model-based methods.  The main
observation is that, when solving linear inverse problems, proximal-based
algorithms separate the operations of measurement consistency and problem
regularization (using prior knowledge). The latter usually consists of a simple
operation, like soft-thresholding, which encodes the assumptions about the
target image and which can be viewed as a denoising step. Given its
independence from the measurements, such operation can be replaced by a more
complex function, such as a CNN. The resulting algorithms are versatile, as the
measurement operator can be easily modified. Most work in this area, however,
has focussed on compressed sensing, in which the measurement operator is
typically a dense random matrix; see~\cite{zhang_ista-net:_2018,
kulkarni_reconnet:_2016, bora_compressed_2017}.

The One-Net~\cite{Chang_OneNet}, for example, replaces the proximal operator
associated to image regularization in an ADMM algorithm with a CNN trained on a
large database. The experiments in~\cite{Chang_OneNet} considered SR, but the
resulting network does not perform as well as current leading CNN-based methods.


The pioneering work in~\cite{DeepImagePriorUlyanov} takes this idea further and
proposes a scheme that requires no training at all. There, an untrained CNN is
used as a prior. Specifically, a linear inverse problem is reparameterized as a
function of the weights of a CNN whose input is a noisy/corrupted image and
whose output is the denoised/reconstructed image. Such reparameterization
provides a type of regularization. 
The work in~\cite{mataev_deepred} combined this idea
with regularization by denoising and used ADMM to solve the resulting linear inverse
problem. See~\cite{IRCNN-zhang2017,HQS} for related work.  
These algorithms require no training at all and can be easily adapted to
different measurement operators. However, they can be particularly slow, as each iteration requires some
backpropagation iterations on the CNN. And, when applied to SR, they are still
outperformed by training-based CNN architectures.

%
         

\section{Proposed Framework} 
\label{Section3}

\subsection{Main Model and Assumptions}
\label{SubSec:ModelAndAssumptions}

We aim to reconstruct the vectorized version of an HR image $x^{\star} \in
\mathbb{R}^n$ from a LR image $b \in \mathbb{R}^m$, with $m<n$. We assume that
these quantities are linearly related:
\begin{equation}
  \label{eq:b}
  b = Ax^{\star},
\end{equation}
where $A \in \mathbb{R}^{m \times n}$ represents the downsampling operator. 
The model in (\ref{eq:b}) is often
used in reconstruction-based and dictionary learning algorithms \cite{aly_image_2005,yang_image_2010,
wavelet_Mallat_2010}, even though many
methods also consider additive noise: $b = Ax^{\star} + \epsilon$,
where $\epsilon$ is a Gaussian random vector \cite{shi_lrtv:_2015-1,
li_self-learning_2016, peyre_non-local_2008, becker_nesta:_2011, Chang_OneNet}.

More interesting, however, is that CNN-based methods implicitly assume the
model in (\ref{eq:b}), although that is rarely acknowledged. In particular, all
the SR networks we know of (e.g., \cite{dong_learning_2014,lai_deep_2017,dong_accelerating_2016,wang2018esrgan}) are trained with HR images that are
downsampled according to \eqref{eq:b}, where $A$ implements bicubic
downsampling. We next discuss other possible choices for $A$. 

\mypar{Common choices for $\boldsymbol{A}$} Different instances of $A \in
\mathbb{R}^{m\times n}$ in \eqref{eq:b} have been assumed in the SR literature:
\begin{itemize}
  \item \textit{Simple subsampling}: $A$ contains equispaced rows of the identity
    matrix $I_n \in \mathbb{R}^{n\times n}$, i.e., each row of $A$ is
    a canonical vector $(0, ..., 0, 1, 0, ...,0)$. This operator is simple to
    implement, but often introduces aliasing.

  \item \textit{Bicubic}: $A = S \cdot B$, where $S$ is a
    simple subsampling operator, and $B$ is a bicubic filter. It is the operator of
    choice for processing training data for CNNs.

  \item \textit{Box-averaging}: if the scaling factor is $s$, then each row of $A$
    contains $s^2$ nonzero elements, equal to $1/s^2$, in positions
    corresponding to a neighborhood of a pixel. In other words, box-averaging
    replaces each block of $s \times s$ pixels by their average. Although
    simpler than the bicubic operator, it does not introduce the aliasing that simple
    subsampling does; see, e.g., \cite{Chang_OneNet}. 
\end{itemize}
In our experiments, we will mostly instantiate $A$ as a bicubic operator. The
reason is that most SR CNNs assume this operator during training. Simple
subsampling and box-averaging will be used to illustrate how our
post-processing scheme adds robustness to operator mismatch, i.e., when $A$ is
different during training and testing.

\begin{figure}[t]
\centering
\includegraphics[width=8.4cm,height=4cm]{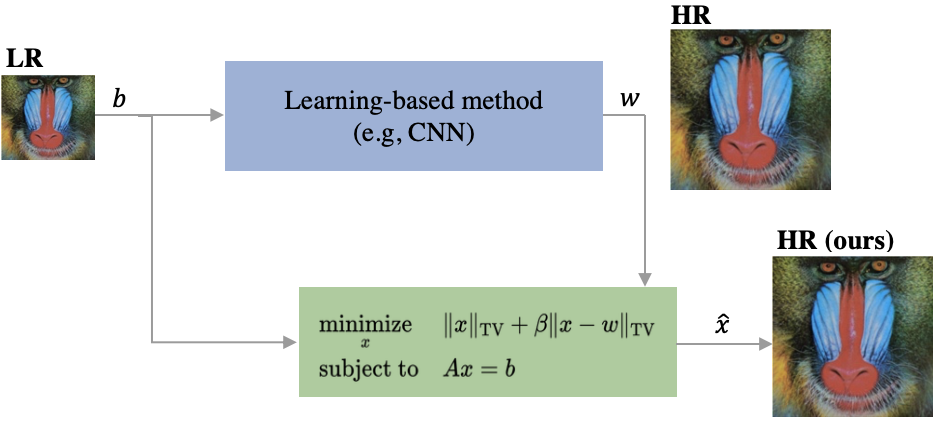}
\caption{
  \footnotesize
  Our framework: the low-resolution image $b$ and the image $w$ super-resolved
  by a CNN are fed into the TV-TV minimization problem which, in turn,
  obtains a high-resolution image $\widehat{x}$ with better quality.
}
\label{fig_framework}
\end{figure}

\mypar{Assumptions} 
We estimate $x^\star \in \mathbb{R}^n$ from $b \in \mathbb{R}^m$ by taking
into account two possibly conflicting assumptions:
\begin{enumerate}
    \item 
      $x^\star$ has a small number of edges, as captured by a
      small total variation (TV);

    \item
      $x^{\star}$ is also close to the side information $w$, the image
      returned by a learning-based method (CNN), where the notion of distance
      is also measured by TV.
\end{enumerate}
For a given vectorization $x \in \mathbb{R}^n$ of an image $X \in \mathbb{R}^{M
\times N}$, the anisotropic 2D TV (semi-)norm is defined as
\cite{rudin_nonlinear_1992,chambolle_algorithm_2004}
\begin{align}
    \left\|x\right\|_{\text{TV}}
  :&=
    \sum_{i=1}^{M} \sum_{j=1}^{N}\left|v_{i
    j}^{\top} x\right|+\left|h_{i j}^{\top} x\right|
  \label{eq:TVdef}
  \vspace{0.2cm}
  \\
  &=
    \bigg\|
    \begin{bmatrix}
      V \\ H
    \end{bmatrix}
    x
    \bigg\|_1
    \label{Eq:DefVH}
  \vspace{0.2cm}
  \\
  &=
  \|Dx\|_1
    \label{Eq:DefD}
  \,.
\end{align}
In~\eqref{eq:TVdef}, $v_{ij}\in \mathbb{R}^n$ and $h_{ij} \in \mathbb{R}^n$
extract the vertical and horizontal differences at pixel $(i,j)$ of $X$.  By
concatenating $v_{ij}$ (resp.\ $h_{ij}$) as rows of $V \in \mathbb{R}^{n\times
n}$ (resp.\ $H\in\mathbb{R}^{n\times n}$), we obtain the representation
in~\eqref{Eq:DefVH}, where $\|\cdot\|_1$ is the $\ell_1$-norm (sum of absolute
values). And the matrix $D\in \mathbb{R}^{2n \times n}$ in~\eqref{Eq:DefD} is
the vertical concatenation of $V$ and $H$. We assume periodic boundaries, so
that both $V$ and $H$ are circulant. As circulant matrices are diagonalizable
by the DFT, matrix-vector products by both $V$ and $H$ can be computed via the
FFT in $O(n\log n)$ time.

\subsection{Our Framework}

The framework we propose is shown schematically in Fig.~\ref{fig_framework}. It
starts by super-resolving $b$ into $w \in \mathbb{R}^n$ with a base method,
which we assume is implemented by a CNN due to their current outstanding
performance. As explained in Section~\ref{Section1}, CNNs fail to enforce
measurement consistency during testing, i.e., $Aw \neq b$ for any matrix $A$
that is assumed to implement the downsampling operation. 

We propose to use an additional block that takes in both the HR output $w$ of
the CNN and the LR image $b$, and creates another HR image $\widehat{x}$. The
block implements what we call TV-TV minimization, which enforces
measurement consistency while guaranteeing that assumptions 1) and 2) are met.
This is explained next.

\mypar{TV-TV minimization}
Given the LR image $b$ and a HR image $w$, \textit{TV-TV minimization} consists
of
\begin{equation}
  \begin{array}[t]{ll}{
    \underset{x}{\text{minimize}}} & \|x\|_{\text{TV}}+\beta\|x-w\|_{\text{TV}} 
    \\ 
    \text{subject to} & Ax=b\,,
  \end{array}
  \label{eq:TV-TV}
\end{equation}
where $x \in \mathbb{R}^n$ is the optimization variable, and $\beta \geq 0$
tradeoffs between assumptions 1) and 2). Indeed, the first term in the
objective of~\eqref{eq:TV-TV} encodes assumption 1), the second term assumption
2), and the constraints enforce measurement consistency. Of course, our 
assumptions 1)-2) can be easily modified to better capture the class of images
to be super-resolved. We found that using TV semi-norms in the objective
yielded better results. In addition, as these functions are convex, problem~\eqref{eq:TV-TV} is convex as well.

Although a problem like~\eqref{eq:TV-TV} has appeared before
in~\cite{chen_prior_2008} in the context of dynamic computed tomography (CT),
the side information $w$ there was an image reconstructed by solving the same
problem in the previous instant; see
also~\cite{mota_compressed_2017,weizman_reference-based_2016,mota_2016_adaptiverate}.
\textit{Our approach is conceptually different in that we use~\eqref{eq:TV-TV}
to improve the reconstruction of a CNN-based method.} 

Next, we show how TV-TV minimization relates to $\ell_1$-$\ell_1$ minimization,
and how the theory for the latter in~\cite{mota_compressed_2017} suggests that
selecting $\beta=1$ in~\eqref{eq:TV-TV} may lead to better performance. 

\mypar{Relation to \boldmath{$\ell_1$-$\ell_1$} minimization} 
Introducing an auxiliary variable $u \in \mathbb{R}^{2n}$ and defining
$\overline{w} := Dw$, we rewrite~\eqref{eq:TV-TV} as
\begin{equation}
  \label{Eq:TVTVintermediate}
  \begin{array}[t]{ll}
    \underset{u, x}{\text{minimize}} & \|u\|_1 +  \beta \|u-\overline{w}\|_1 \\
    \text{subject to} &  Ax = b \\
    & Dx = u\,.
  \end{array}
\end{equation}
Thus, $\overline{x} := (u, x) \in \mathbb{R}^{3n}$ is the full optimization variable. Define 
$\overline{A} 
:= 
\begin{bmatrix}
  0_{m\times 2n} & A;& -I_{2n} & D
\end{bmatrix}
$, 
and 
$
\overline{b} 
:=
\begin{bmatrix}
    b & 0_{2n}
\end{bmatrix}^\top
$, 
where $0_{a\times b}$ (resp. $0_a$) represents the zero matrix (resp.\ vector)
of dimensions $a \times b$ (resp.\ $a \times 1$), and $I_{2n}$ is the identity
matrix in $\mathbb{R}^{2n}$. This enables us to rewrite
(\ref{Eq:TVTVintermediate}) as 
\begin{equation}
  \label{Eq:L1L1}
  \begin{array}[t]{ll}
    \underset{\overline{x}}{\text{minimize}} & \|G_{2n} \overline{x}\|_1 
    + \beta \|G_{2n} \overline{x} - \overline{w}\|_1\\
    \text{subject to} &  \overline{A}\overline{x} = \overline{b}\,,
  \end{array}
\end{equation}
where $G_{2n} \in \mathbb{R}^{2n \times 3n}$ contains the first $2n$ rows of
the identity matrix $I_{3n}$. In other words, for a vector $v
\in \mathbb{R}^{3n}$, $G_{2n} v$ represents the first $2n$ components of $v$.
The work in \cite{mota_compressed_2017} analyzes \eqref{Eq:L1L1} when $G_{2n}$
is the full identity matrix and the entries of $\overline{A}$ are drawn from a Gaussian
distribution. 
Specifically, it provides the number of measurements required for perfect
reconstruction under these assumptions. It is shown both theoretically and
experimentally that the best reconstruction performance is obtained when $\beta
= 1$. Although the theory in \cite{mota_compressed_2017} cannot be easily
extended\footnote{The reason is that the matrix $\overline{A}$ is very
structured and thus, even when $A$ is assumed Gaussian, its nullspace is not
uniformly distributed.} to \eqref{Eq:L1L1}, our experiments indicate that
$\beta = 1$ still leads to the best results in our setting. 

\subsection{Algorithm for TV-TV minimization} 

We now explain how to efficiently solve TV-TV minimization~\eqref{eq:TV-TV}
with the alternating direction method of multipliers
(ADMM)~\cite{boyd_distributed_2011}.  In contrast with the majority of SR
algorithms, which operate on individual patches, our algorithm operates on full
images. We do that by capitalizing on the fact that matrix-vector
multiplications can be performed fast whenever the matrix is $D$
[cf.~\eqref{eq:TVdef}] or any of the instantiations of $A$ mentioned in
Section~\ref{SubSec:ModelAndAssumptions}. 

\mypar{ADMM} 
The problem that ADMM solves is
 \begin{equation}
   \label{Eq:ProbSolvedByADMM}
   \begin{array}[t]{ll}
     \underset{y, z}{\text{minimize}}  & f(y) + g(z)
     \\
     \text{subject to} & F y + G z = 0\,,
  \end{array}
\end{equation}
where $f$ and $g$ are closed, proper, and convex functions, and
$F$ and $G$ are given matrices. Associating a dual variable $\lambda$ to
the constraints of~\eqref{Eq:ProbSolvedByADMM}, ADMM iterates on $k$
\begin{subequations}\label{Eq:ADMM}
  \begin{align}
    y^{k+1} 
    &= 
    \underset{y}{\text{argmin}}\,\, 
    f(y) + \frac{\rho}{2}\big\|F y + G z^k + \lambda^k\big\|_2^2 
    \label{Eq:ADMMy}
    \\
    z^{k+1} 
    &= 
    \underset{z}{\text{argmin}}\,\, 
    g(z) + \frac{\rho}{2}\big\|F y^{k+1} + Gz  + \lambda^k\big\|_2^2 
    \label{Eq:ADMMz}
    \\
    \lambda^{k+1}
    &=
    \lambda^{k} + F y^{k+1} + G z^{k+1}
    \label{Eq:ADMMLambda}
    \,,
\end{align}
\end{subequations}
where $\rho > 0$ is the augmented Lagrangian parameter.

\mypar{Applying ADMM} 
Although there are many possible reformulations of~\eqref{eq:TV-TV} to which
ADMM is applicable, they can yield different performances. Our
reformulation simply adds another variable $v \in \mathbb{R}^n$
to~\eqref{Eq:TVTVintermediate} [which is equivalent to~\eqref{eq:TV-TV}]:
\begin{equation}
  \label{eq:admm_reform}
  \begin{array}[t]{ll}
    \underset{u,x,v}{\text{minimize}} & \|u\|_1 +  \beta \|u-\overline{w}\|_1\\
    \text{subject to} &  Ax = b\\ & Dv=u\\ & v=x\,.
  \end{array}
\end{equation}
We establish the following correspondence between~\eqref{Eq:ProbSolvedByADMM}
and~\eqref{eq:admm_reform}: we set $y = (u,x)$, $z =
v$, and assign
\begin{align*}
  f(u,x) &= \|u\|_1 + \beta\|u - \overline{w}\|_1 + \text{i}_{Ax = b}(x)
  &
  F &= 
  \begin{bmatrix}
    -I_{2n} & 0
    \\
    0 & I_{n}
  \end{bmatrix}
  \\
  g(v) &= 0
  &
  G &=
  \begin{bmatrix}
    D
    \\
    -I_n
  \end{bmatrix}\,,
\end{align*}
where $\text{i}_{Ax = b}(x)$ is the indicator function of $Ax = b$, i.e., it
evaluates to $0$ if $Ax = b$, and to $+\infty$ otherwise. This means that we
dualize only the last two constraints of~\eqref{eq:admm_reform}, and thus
$\lambda$ has two components: $\lambda = (\eta, \mu) \in \mathbb{R}^{2n} \times
\mathbb{R}^{n}$. The above correspondence yields closed-form solutions for the
problems in~\eqref{Eq:ADMMy} and~\eqref{Eq:ADMMz} (see below). Furthermore,
even though the objective of~\eqref{eq:admm_reform} is not strictly convex, the
fact that $F$ and $G$ have full column-rank implies that the sequence $(y^k,
z^k)$ generated by ADMM~\eqref{Eq:ADMM} has a \textit{unique} limit point,
which solves~\eqref{eq:admm_reform} \cite{Mota11-ADMMProof}.  We now elaborate on
how to solve~\eqref{Eq:ADMMy}-\eqref{Eq:ADMMz}.

\mypar{Solving~\eqref{Eq:ADMMy}} 
Using the above correspondence, problem~\eqref{Eq:ADMMy} decouples into two
independent problems that can be solved in parallel:
\begin{align}
  u^{k+1}
  &=
  \underset{u}{\text{argmin}} \,\,\,  
  \|u\|_1 +  \beta \|u-\overline{w}\|_1 
  + 
  \frac{\rho}{2}\|u - s^k\|_2^2
  \label{Eq:ADMMSubProbU}
  \\
  x^{k+1}
  &=
  \begin{array}[t]{cl}
    \underset{x}{\text{argmin}} & \frac{1}{2}\|x-p^k\|^2_2\\
    \text{s.t.} &  A x = b\,,
  \end{array}
  \label{Eq:ADMMSubProbX}
\end{align}
where we defined $s^k := Dv^k + \eta^{k}$ and $p^k := v^k - \mu^k$.

Problem~\eqref{Eq:ADMMSubProbU} decomposes further componentwise, and the
solution for each component can be obtained by evaluating the respective optimality
condition (via subgradient calculus). Namely, for $i = 1, \ldots, 2n$, if
$\overline{w}_i \geq 0$, then component $u_i^{k+1}$ is given by
\begin{equation}
  \label{Eq:ComponentwiseSolutionUPositive}
  \left\{
    \begin{array}{ll}
      s_i - \frac{1}{\rho}(\beta + 1) 
      &,\,\, 
      s_i > \overline{w}_i + \frac{1}{\rho}(\beta + 1)
      \vspace{0.1cm}
      \\
      \overline{w}_i 
      &,\,\, 
      \overline{w}_i - \frac{1}{\rho}(\beta - 1) 
      \leq 
      s_i 
      \leq 
      \overline{w}_i + \frac{1}{\rho}(\beta + 1) 
      \vspace{0.1cm}
      \\
      s_i + \frac{1}{\rho}(\beta - 1)
      &,\,\, 
      -\frac{1}{\rho}(\beta - 1) < s_i < \overline{w}_i - \frac{1}{\rho}(\beta - 1)
      \vspace{0.1cm}
      \\
      0  
      &,\,\, 
      -\frac{1}{\rho}(\beta - 1) \leq s_i \leq -\frac{1}{\rho}(\beta - 1)
      \vspace{0.1cm}
      \\
      s_i + \frac{1}{\rho}(\beta + 1)
      &,\,\, 
      s_1 < -\frac{1}{\rho}(\beta + 1)\,.
    \end{array}
  \right.
\end{equation}
And if $\overline{w}_i < 0$, then component $u_i^{k+1}$ is given by
\begin{equation}
  \label{Eq:ComponentwiseSolutionUNegative}
  \left\{
    \begin{array}{ll}
      s_i + \frac{1}{\rho}(\beta + 1) 
      &,\,\, 
      s_i > \frac{1}{\rho}(\beta + 1)
      \vspace{0.1cm}
      \\
      0                                
      &,\,\, 
      \frac{1}{\rho}(\beta - 1) \leq s_i \leq \frac{1}{\rho}(\beta + 1)
      \vspace{0.1cm}
      \\
      s_i - \frac{1}{\rho}(\beta - 1)
      &,\,\, 
      \overline{w}_i + \frac{1}{\rho}(\beta - 1) < s_i < \frac{1}{\rho}(\beta - 1)
      \vspace{0.1cm}
      \\
      \overline{w}_i
      &,\,\, 
      \overline{w}_i - \frac{1}{\rho}(\beta + 1)
      \leq
      s_i
      \leq
      \overline{w}_i + \frac{1}{\rho}(\beta - 1)
      \vspace{0.1cm}
      \\
      s_i + \frac{1}{\rho}(\beta + 1)
      &,\,\, 
      s_i < \overline{w}_i - \frac{1}{\rho}(\beta + 1)\,.
    \end{array}
  \right.
\end{equation}

Problem~\eqref{Eq:ADMMSubProbX} is the projection of $p^k$ onto the solutions
of $Ax = b$. Assuming that $A$ has full row-rank, i.e., $AA^\top$ is
invertible, \eqref{Eq:ADMMSubProbX} also has a closed-form solution:
\begin{equation}
    x^{k+1} = p^k - A^\top(AA^\top)^{-1}(Ap-b)\,,
    \label{eq:x3}
\end{equation}
whose computation has a complexity that depends on the properties of the
downsampling operator $A$.  When $A$ is simple subsampling or the box-averaging
operator, $A A^\top$ is the identity matrix $I_m$ or a multiple of it. In that
case, computing~\eqref{eq:x3} requires only two matrix-vector operations which,
due to the structure of $A$, can be implemented by indexing. In other words,
there is no need to construct $A$ explicitly.

On the other hand, when $A$ is the bicubic operator, the inverse of $A A^\top$
can no longer be computed easily, and we solve the linear system
in~\eqref{eq:x3} with the conjugate gradient method. In this case,
matrix-vector products can be computed in $O(n\log n)$ time using the FFT.

\mypar{Solving~\eqref{Eq:ADMMz}} 
With our choice of $g$, $F$, and $G$, problem~\eqref{Eq:ADMMz} becomes
\begin{align}
  v^{k+1} 
  &= 
  \underset{v}{\text{argmin}} \,\,\,
  \frac{1}{2}\big\|Dv - u^{k+1} + \eta^k\big\|_2^2 + \frac{1}{2}\big\|v -
  x^{k+1} + \mu^k\big\|_2^2
  \notag
  \\
  &=  
  (I_n + D^\top D)^{-1} \Big[x^{k+1} - \mu^k + D^\top (u^{k+1} - \eta^k)\Big]\,.
  \label{eq:v}
\end{align}
Given the definition of $D$ in~\eqref{Eq:DefVH}-\eqref{Eq:DefD}, 
we have 
\begin{align*}
  I_n + D^\top D 
  &= 
  I_n + V^\top V + H^\top H
  \\
  &=
  C_n^H \Big(I_n + \text{Diag}(C_n v)^2 + \text{Diag}(C_n h)^2\Big)C_n\,,
\end{align*}
where the last step uses the fact that $V$ and $H$ are circulant matrices and,
therefore, are generated by some vectors $v$ and $h$, respectively. Also, $C_n$
denotes the DFT matrix in $\mathbb{R}^n$, and $\text{Diag}(x)$ is a diagonal
matrix with the entries of $x$ in its diagonal. This representation of $I_n +
D^\top D$ not only enables us to compute its inverse in closed-form (just take
the inverse of the matrix in parenthesis), but also to do it without
constructing any matrix explicitly.

\mypar{Dual updates}
Finally, since $\lambda$ decomposes as $(\eta, \mu)$, the dual variable update
in~\eqref{Eq:ADMMLambda} becomes
\begin{subequations}
  \label{Eq:DualVariableUpdates}
\begin{align}
  \eta^{k+1} &= \eta^k + D v^{k+1} - u^{k+1}
  \label{SubEq:DualVariableUpdatesEta}
  \\
  \mu^{k+1} &= \mu^k + x^{k+1} - v^{k+1}\,.
  \label{SubEq:DualVariableUpdatesMu}
\end{align}
\end{subequations}
Applying ADMM~\eqref{Eq:ADMM} to the equivalent
reformulation~\eqref{eq:admm_reform} of TV-TV minimization~\eqref{eq:TV-TV}
therefore yields
steps~\eqref{Eq:ComponentwiseSolutionUPositive}-\eqref{Eq:ComponentwiseSolutionUNegative}
for each component of $u$, \eqref{eq:x3} for $x$, \eqref{eq:v} for $v$,
and~\eqref{Eq:DualVariableUpdates} for the dual variables. These steps are
repeated iteratively until a stopping criterion is met; we use the one suggested in
\cite{boyd_distributed_2011}.

\section{Experiments} 
\label{Section4}

We now describe our experiments. After explaining the experimental setup, we
expand on the phenomenon described in Fig.~\ref{fig_prob}. Then, we consider
the case of operator mismatches (i.e., $A$ is different during training and
testing), and show how our framework adds significant robustness in this
scenario. Finally, we report experiments on standard SR datasets.
Code to replicate our experiments is available online.\footnote{https://github.com/marijavella/sr-via-CNNs-and-tvtv} 

\subsection{Experimental Setup}

\mypar{Algorithm parameters} 
Most experiments were run using the same algorithm settings, unless indicated
otherwise.  The hyperparameter $\beta$ in~\eqref{eq:TV-TV} was always set to
$1$ and, for most experiments, $A$ was the bicubic operator via MATLAB's
\textsc{imresize}.  For ADMM, we adopted the stopping criterion
in~\cite[\S3.3.1]{boyd_distributed_2011} with $\epsilon^{\text{pri}} =
\epsilon^{\text{dual}} = 0.001$, or stopped after $1500$ iterations. Also, we
initialized $\rho = 0.5$ and adjusted it automatically using the heuristic
in~\cite[\S3.4.1]{boyd_distributed_2011}. 


\mypar{Datasets} 
We considered the standard SR test sets Set5
\cite{bevilacqua_low-complexity_2012}, Set14 \cite{zeyde_single_2012}, BSD100
\cite{martin_database_2001} and Urban100 \cite{huang_single_2015}, which
contain images of animals, buildings, people, and landscapes. 

\mypar{Computational platform} 
All experiments were run on Matlab (R2019a) using a workstation with 12 core
2.10GHz Intel(R) Xeon(R) Silver 4110 CPU and two Nvidia GeForce RTX GPUs. 

\mypar{Methods evaluated} 
We compared our framework against the state-of-the-art methods in
Table~\ref{tab:summary} and also considered simple TV minimization, i.e.,
\eqref{eq:TV-TV} with $\beta=0$, using the TVAL3 solver
\cite{li_efficient_2013}. The table shows the acronyms and references of the
methods, their main technique, the scaling factors (S.F.) considered in the
original papers, and the datasets used for training. Note that all methods except
LapSRN and ESRGAN were evaluated only for $2 \times$ and $4 \times$ scaling
factors. LapSRN also handles an $8 \times$ scaling factor, while ESRGAN only
handles $4 \times$. 
The training datasets in Table~\ref{tab:summary} have 
91 (T91),
100 (General100), 
200 (BSDS200), 
324 (OutdoorSceneTraining),
500 (BSDS500), 
800 (DIV2K), 
2650 (Flickr2K), 
4744 (WED), 
and 396,000 (ImageNet) images. 

\isdraft{
\begin{table}
\centering
\caption{\MakeUppercase{Methods used in our experiments. For each, we show the main technique, the
scaling factors it can handle and, if any, the training dataset.} }
  \label{tab:summary}
  \includegraphics[scale=0.95]{images/TabI.pdf}
\end{table}
}
{
\begin{table}
	\centering
 \renewcommand{\arraystretch}{1.2}
\caption{
\MakeUppercase{Methods used in our experiments. For each, we show the main technique, the
scaling factors it can handle and, if any, the training dataset.}
}
  \vspace{0.15cm}
\begin{tabular}{@{}llll@{}}\toprule
  \textcolor{thisblue}{\sf\textbf{Method}} 
  & 	
  \textcolor{thisblue}{\sf\textbf{Type}} 
  & 	
  \textcolor{thisblue}{\sf\textbf{S.F.}} 
  & 
  \textcolor{thisblue}{\sf\textbf{Training dataset}} \\
  \midrule
  \textcolor{darkgreen}{\sf Kim} \sf\cite{kim_single-image_2010} 
  & 
  \sf Regression & \sf 2, 4 &\\
  \midrule
  \textcolor{darkgreen}{\sf SRCNN} \sf \cite{dong_learning_2014} 
  & 
  \sf CNN & \sf 2, 4  & \sf ImageNet \sf\cite{Deng09imagenet}\\
  \midrule
  \textcolor{darkgreen}{\sf SelfExSR} \sf\cite{huang_single_2015} 
  & 
  \sf Self-similarity 
  & \sf 2, 4  & \\
  \midrule
  \textcolor{darkgreen}{\sf FSRCNN} \sf\cite{dong_accelerating_2016}
  & \sf CNN & \sf 2, 4 & \sf T91 \cite{yang_image_2010}, General100 \cite{dong_accelerating_2016}  \\
  \midrule
  \textcolor{darkgreen}{\sf DRCN} \sf\cite{kim_DRCN} 
  & 
  \sf CNN & \sf 2, 4  & \sf T91 \cite{yang_image_2010} \\
  \midrule
  \textcolor{darkgreen}{\sf VDSR} \sf\cite{Kim_VDSR} 
  & 
  \sf CNN & \sf 2, 4  & \sf T91 \cite{yang_image_2010}, BSDS200 \cite{BSDS200} \\
  \midrule
  \textcolor{darkgreen}{\sf LapSRN} \sf\cite{lai_deep_2017} 
  & 
  \sf CNN & \sf 2, 4, 8  & \sf T91 \cite{yang_image_2010}, BSDS200 \cite{BSDS200} \\
  \midrule
  \multirow{2}{*}{\textcolor{darkgreen}{\sf SRMD} \sf\cite{zhang_learning_2018}} 
  &   
  \multirow{2}{*}{\sf CNN} 
  & 
  \multirow{2}{*}{\sf 2, 4} & \sf  DIV2K \cite{Div2k}, BSDS200 \cite{BSDS200} 
  \\
  & & & \sf WED \cite{WED} 
  \\
  \midrule
  \multirow{2}{*}{\textcolor{darkgreen}{\sf IRCNN} \sf\cite{IRCNN-zhang2017} } 
  & \multirow{2}{*}{\sf Plug-and-play}  &  \multirow{2}{*}{\sf 2, 4} & \sf ImageNet \cite{Deng09imagenet}, WED \cite{WED},  
  \\
  & & & \sf BSDS500 \cite{BSDS500} \\
  \midrule
  \multirow{2}{*}{\textcolor{darkgreen}{\sf ESRGAN} \sf\cite{wang2018esrgan}} 
  &  
  \multirow{2}{*}{\sf GAN}  &  \multirow{2}{*}{\sf 4} & \sf DIV2K \cite{Div2k}, Flickr2K \cite{flick2r},  \\
  &  &  & \sf OutdoorSceneTraining \cite{Outdoorscene-dataset}\\
  \midrule
  \textcolor{darkgreen}{\sf DeepRED} \sf\cite{DeepImagePriorUlyanov}
  & 
  \sf Plug-and-play 
  & \sf 2, 4 & \\
  \midrule
  \textcolor{darkgreen}{ \sf TVAL3} \sf\cite{li_efficient_2013}
  & \sf Optimization & \sf 2, 4 &   \\
  \bottomrule
\end{tabular}
\label{tab:summary}
\end{table}
}

Both during training and testing, all the CNN-based methods in
Table~\ref{tab:summary} extract the luminance channel of the YCbCr color space,
and then convert the image from \textsc{uint8} to \textsc{double}. During
training, the HR images are converted to LR images by applying MATLAB's
\textsc{imresize}, as originally done in \cite{dong_learning_2014}. Other
methods, such as~\cite{yulun_RCAN}, work on \textsc{uint8} images directly or
use Python's \textsc{imresize}, which is different from MATLAB's. To keep our
experiments consistent, we did not consider such methods. 

The output images for Kim \cite{kim_single-image_2010}, SRCNN
\cite{dong_learning_2014} and SelfExSR \cite{huang_single_2015} were retrieved
from an online repository.\footnote{https://github.com/jbhuang0604/SelfExSR}
For the remaining methods, we generated the outputs from the available
pretrained models. 

\mypar{Performance metrics} 
We compared different algorithms by evaluating the PSNR (dB) and SSIM
\cite{wang_image_2004} on the luminance channel of the output images. As these
metrics do not often capture perceptual quality, we also provide sample images
for qualitative evaluation. 

\subsection{Measurement Inconsistency of CNNs}
\label{SubSec:ExpMeasurementInconsistency}

We show that the phenomenon illustrated in Fig.~\ref{fig_prob} for
SRCNN~\cite{dong_learning_2014} occurs not only for this network, but is
pervasive. That is, CNNs for SR fail to enforce measurement
consistency~\eqref{eq:b} during testing. We chose three images for this
purpose: \textit{Baboon} from Set14 \cite{zeyde_single_2012}, \textit{38092}
from BSD100, and \textit{img}$_{005}$ from Urban100 \cite{huang_single_2015}.
Every image is downsampled with MATLAB's \textsc{imresize}, which is the
procedure executed for training each CNN, and the resulting LR image is fed
into the network. We chose a scaling factor of $4$.

\mypar{Results}
Table~\ref{tab:1} shows the results for a subset of methods in
Table~\ref{tab:summary}. In the 3rd column, it displays the $\ell_2$-norm of the difference
between the downsampled HR outputs, i.e., $Aw$, and the input LR image $b$; in the 4th column, it shows the same quantity after feeding the corresponding $w$ (and $b$, cf.\
Fig.~\ref{fig_framework}) to our method. It
can be seen our post-processing improves consistency by 6 orders of magnitude.  Note
that even though SRMD models various degradations without retraining, it still
fails to ensure consistency. IRCNN is a plug-and-play method and, as a result,
can also handle different degradation models. Although it achieves better
consistency than pure CNN-based methods, it is still 5 orders of magnitude
below our scheme. The last row of Table~\ref{tab:1} shows the consistency of
DeepRED~\cite{mataev_deepred}, which processes the three RGB channels
simultaneously. For this reason, it was difficult to provide a fair comparison
with our method.

\isdraft{
\begin{table}
\centering
  	\caption{\MakeUppercase{Consistency achieved by CNN-type methods} ($\|Aw - b\|_2$) \MakeUppercase{and by our algorithm} ($\|A\hat{x} - b\|_2$).}
  \label{tab:1}
  \includegraphics[scale=0.95]{images/TabII.pdf}
\end{table}
}
{
\begin{table}
\sffamily\selectfont
	\centering
 \renewcommand{\arraystretch}{1.2}
\caption{
 \MakeUppercase{Consistency achieved by CNN-type methods} ($\|Aw - b\|_2$) \MakeUppercase{and by our
  algorithm} ($\|A\hat{x} - b\|_2$).
\vspace{0.15cm}
  \sffamily
}
    \begin{tabular}{@{}llll@{}}\toprule
	\textcolor{thisblue}{\textbf{Method}} & 	\textcolor{thisblue}{\textbf{Image}} & 	\textcolor{thisblue}{$\boldsymbol{\|Aw-b\|_2}$} & \textcolor{darkredc}{$\boldsymbol{\|A\hat{x}-b\|_2}$} \\
\midrule
	\multirow{3}{*} {\textcolor{darkgreen}{SRCNN} \cite{dong_learning_2014}} & \textit{Baboon} & $\phantom{0}5.29\times10^{-1}$  & $\boldsymbol{4.72\times10^{-7}}$\\
               & \textit{38092} & $\phantom{0}5.68\times10^{-1}$  & {$\boldsymbol{4.18\times10^{-7}}$}\\
               & \textit{img$_{005}$} & $14.93\times10^{-1}$ & $\boldsymbol{7.72\times10^{-7}}$\\
\midrule
	\multirow{3}{*}{\textcolor{darkgreen}{FSRCNN} \cite{dong_accelerating_2016}}& \textit{Baboon}  & $\phantom{0}3.26\times10^{-1}$  &  $\boldsymbol{4.81\times10^{-7}}$\\
                & \textit{38092}  & $\phantom{0}2.91\times10^{-1}$ & {$\boldsymbol{3.80\times10^{-7}}$}  \\
                &\textit{img$_{005}$}& $10.32\times10^{-1}$  & $\boldsymbol{3.55\times10^{-7}}$ \\
\midrule
	\multirow{3}{*}{\textcolor{darkgreen}{SRMD} \cite{zhang_learning_2018}} &   \textit{Baboon}   & $\phantom{0}4.14\times10^{-1}$  & $\boldsymbol{6.62\times10^{-7}}$\\
                         &  \textit{38092} & $\phantom{0}2.39\times10^{-1}$  & $\boldsymbol{9.48\times10^{-7}}$ \\ 
                         &\textit{img$_{005}$}& $\phantom{0}9.53\times10^{-1}$ &  $\boldsymbol{9.89\times10^{-7}}$ \\
                         \midrule
	\multirow{3}{*}{\textcolor{darkgreen}{IRCNN} \cite{IRCNN-zhang2017}} &   \textit{Baboon}   &$\phantom{0}8.72\times10^{-2}$ & $\boldsymbol{5.42\times10^{-7}}$ \\
                         & \textit{38092} & $\phantom{0}8.06\times10^{-2}$ &  $\boldsymbol{4.08\times10^{-7}}$\\ 
                         &\textit{img$_{005}$}& $\phantom{0}4.57\times10^{-1}$ & $\boldsymbol{6.70\times10^{-7}}$ \\
                         \midrule
	\multirow{3}{*}{\textcolor{darkgreen}{DeepRed} \cite{mataev_deepred}} &   \textit{Baboon}   & $\phantom{0}2.62\times10^{-1}$&  \\
                         & \textit{38092} & $\phantom{0}2.06\times10^{-1}$ &  \\ 
                         &\textit{img$_{005}$}&$\phantom{0}3.11\times10^{-1}$ &  \\
\bottomrule
\end{tabular}
\label{tab:1}
\end{table}
}

\isdraft{
\begin{table}
\centering
  	\caption{OPERATOR MISMATCH EXPERIMENTS. PSNR VALUES UNDER DIFFERENT SAMPLING OPERATORS 
  FOR $A$: BICUBIC, BOX FILTERING, AND SIMPLE SUBSAMPLING. WITHIN EACH BOX, THE BEST (HIGHER) VALUES ARE HIGHLIGHTED IN \textbf{BOLD}.}
  \label{tab:2}
  \includegraphics[scale=0.15]{images/TabIII.jpg}
\end{table}
}
{
\begin{table}
\sffamily\selectfont
\centering
\renewcommand{\arraystretch}{1.2}
\caption{
  \MakeUppercase{
  Operator mismatch experiments. PSNR values under different sampling operators
  for $A$: bicubic, box filtering, and simple subsampling. Within each box, the
  best (higher) values are highlighted in \textbf{bold}.}
}
  \vspace{0.15cm}
\resizebox{\columnwidth}{!}{%
  \begin{tikzpicture}[overlay]
    \begin{scope}[on background layer]
      \fill[black!12!](3.3,-3.201) rectangle (5.3,3.35);
      \fill[black!12!](5.4,-3.201) rectangle (7.4,3.35);
      \fill[black!12!](7.5,-3.201) rectangle (9.49,3.35);
    \end{scope} 
  \end{tikzpicture}
\begin{tabular}{ @{}llllllll@{} }
	\toprule
		\textcolor{thisblue} {\textbf{Method}} &  	\textcolor{thisblue}
    {\textbf{Image}} & 	\textcolor{thisblue} {{\textbf{Bic.}}} &
    \textcolor{darkredc}{\sf \textbf{\textit{Ours}}}  &
    \textcolor{thisblue}{\textbf{Box}} &  \textcolor{darkredc}{\sf
    \textbf{\textit{Ours}}}  & 	\textcolor{thisblue}{\textbf{Sub.}} &
    \textcolor{darkredc} {\sf \textbf{\textit{Ours}}}  \\
	\midrule
    \multirow{3}{*} {\textcolor{darkgreen}{\sf SRCNN} \cite{dong_learning_2014}} & \textit{Baboon} & 22.70 & \textbf{22.73} & 22.49 &\textbf{22.53} & 17.48& \textbf{19.31}\\
	&  \textit{38092} & 25.90  & \textbf{25.95} & 25.69 & \textbf{25.77} & 20.07 & \textbf{21.98 }\\
	& \textit{img$_{005}$} & 25.12 & \textbf{25.28} & 24.99  & \textbf{25.24} & 17.92 & \textbf{19.91}\\
	\hline
	\multirow{3}{*}  {\textcolor{darkgreen}{FSRCNN} \cite{dong_accelerating_2016}} & \textit{Baboon} & 22.79 & \textbf{22.80 }& 22.49 & \textbf{22.56} & 17.38 & \textbf{19.45}\\
	&  \textit{38092} & 26.03 &  \textbf{26.06} & 25.64 & \textbf{25.74} & 20.00 & \textbf{22.05}\\
	& img$_{005}$ & 25.81 & \textbf{25.87} & 25.12 & \textbf{25.38}& 17.79 & \textbf{19.85}\\
	\hline
	\multirow{3}{*}   {\textcolor{darkgreen}{SRMD} \cite{zhang_learning_2018}} &  \textit{Baboon} & 22.90  & \textbf{22.91} & 22.52 &\textbf{22.61} & 16.95 & \textbf{19.21}  \\
	&  \textit{38092} & 26.20 & \textbf{26.21}  & 25.62 & \textbf{25.74 }& 19.39 & \textbf{21.70}\\ 
	& img$_{005}$ & 26.56 & \textbf{26.60} &  25.59 & \textbf{25.96} & 17.36 & \textbf{19.58 }\\
	\hline
	\multirow{3}{*}  {\textcolor{darkgreen}{IRCNN} \cite{IRCNN-zhang2017}} & \textit{Baboon} & \textbf{22.76}  & \textbf{22.76} & \textbf{22.51} & \textbf{22.51}  & 17.41 & \textbf{19.59} \\
	&  \textit{38092} & \textbf{26.09} & \textbf{26.09}  & 25.77  & \textbf{25.78} & 20.54 & \textbf{22.53} \\ 
	& img$_{005}$ & 26.18  & \textbf{26.20}  & \textbf{25.86} & \textbf{25.86} & 18.23 & \textbf{19.83}\\
	\hline
	\multirow{3}{*}  {\textcolor{darkgreen}{TVAL3} \cite{li_efficient_2013}} & \textit{Baboon} & 22.40  &  & 22.27 &  & 20.83 &  \\
	&  \textit{38092} & 25.59&  &  25.00 & & 21.35 &  \\ 
	& \textit{img$_{005}$} & 24.29 &  & 22.54 &  & 17.28 & \\
	\bottomrule
\end{tabular}
}
\label{tab:2}
\end{table}
}

\subsection{Robustness to Operator Mismatch}

As previously stated, most SR CNNs are trained by downsampling a HR into a LR
image using the bicubic operator. If, during testing, $A$ is different from the
bicubic operator then, as we will see, there can be a serious drop in
performance. This may indeed limit the applicability of CNNs in real-life
scenarios. Our approach, however, mitigates this effect and adds robustness to
the SR task. 
We considered the same images and methods as in Table~\ref{tab:1}, with DeepRed
replaced by TVAL3, and considered the operators for $A$ described in
Section~\ref{SubSec:ModelAndAssumptions}: bicubic, box averaging, and simple
subsampling. 

\mypar{Results}
Each shaded box in Table~\ref{tab:2} shows, for each subsampling operator, the
PSNR values obtained by a given method, and by subsequently processing its
output with our scheme. While all methods perform the best under bicubic
subsampling, there is a performance drop for box filtering, and an even larger
drop for simple subsampling. Note that our method systematically improves the
output of all the networks, even for bicubic subsampling. And while the
improvement is of less than 1dB for bicubic subsampling, it averages around
2dBs for simple subsampling. Indeed, the performance of the CNNs for this case
drops so much that there is a large margin for improvement. Interestingly,
TVAL3, which solves~\eqref{eq:TV-TV} with $\beta = 0$, is the worst method for
bicubic subsampling, but approaches the performance of CNNs for box averaging
and, besides ours, becomes the best for simple subsampling. Hence, this 
illustrates that reconstruction-based methods can be more robust and
adaptable than CNN architectures. 
\isdraft{
\begin{table*}[t]
\centering
  	\caption{ AVERAGE PSNR (SSIM) RESULTS in dB AND EXECUTION TIME IN SECONDS OF OUR METHOD USING THE REFERENCE METHODS.}
  \label{tab:3}
  \includegraphics[scale=0.95]{images/TabIV.pdf}
\end{table*}
}
{
\begin{table*}[t]
\sffamily\selectfont
    
\begin{tikzpicture}[overlay]
\begin{scope}[on background layer]
\fill[black!12!](5.4,-5.59) rectangle (11.35,-1.4);
\fill[black!12!](11.6,-5.59) rectangle (17.6,-1.4);
\fill[black!12!](5.4,-5.77) rectangle (11.35,-10);
\fill[black!12!](11.6,-5.77) rectangle (17.6,-10);
\fill[black!12!](5.4,-10.14) rectangle (11.35,-13.48);
\fill[black!12!](11.6,-10.14) rectangle (17.6,-13.48);
\fill[black!12!](5.4,-13.6) rectangle (11.35,-19.3);
\fill[black!12!](11.6,-13.6) rectangle (17.6,-19.3);
\fill[black!12!](5.4,-19.45) rectangle (11.35,-23.49);
\fill[black!12!](11.6,-19.45) rectangle (17.6,-23.49);
\end{scope} 
\end{tikzpicture}

\centering
	\renewcommand{\arraystretch}{1.2}
	\caption{ \MakeUppercase{Average PSNR (SSIM) results in} {dB}  \MakeUppercase{and execution time in seconds of our method using the reference methods.}}
	\vspace{0.3cm}
	\label{Tab:SystematicExp}
	\def\tbsp{0.15cm}
	\scalebox{0.97}{
		\begin{tabular}{@{}p{1.2cm}p{0.8cm}p{2.2cm}p{2.2cm}p{2.2cm}p{1cm}p{2.2cm}p{2.2cm}p{1cm}@{}}
			\toprule[1.1pt]
			\textcolor{thisblue}{\sf \textbf{Dataset}} & 
			\textcolor{thisblue}{\sf \textbf{Scale}} & 
			\textcolor{thisblue}{\sf \textbf{TVAL3}} \cite{li_efficient_2013} &
			\textcolor{darkgreen}{\sf Kim}~\cite{kim_single-image_2010} &
			\textcolor{darkredc}{\sf \textit{\textbf{Ours}}}   &
			\textcolor{thisblue}{\sf \textbf{Time}}  &
			\textcolor{darkgreen}{\sf SRCNN}~\cite{dong_learning_2014}  &
			\textcolor{darkredc}{\sf \textit{\textbf{Ours}}}   &
			\textcolor{thisblue}{\sf \textbf{Time}} 
			\\
			\midrule
\multirow{2}{*} {Set5} & $\times2$ & 34.0315 (0.9354)&36.2465 (0.9516)  & \textbf{36.4499} (\textbf{0.9537}) & $\phantom{0}$61.89 & 36.2772 (0.9509) & \textbf{36.5288} (\textbf{0.9536}) &  $\phantom{0}$56.60 \\
			& $\times4$ & 29.1708 (0.8349) & 30.0730 (0.8553)& \textbf{30.2289} (\textbf{0.8593}) &  $\phantom{0}$46.96 &30.0765 (0.8525)  & \textbf{30.2669} (\textbf{0.8590}) &  $\phantom{0}$46.93 \\
			\midrule
			\multirow{2}{*}{Set14} & $\times2$ & 31.0033 (0.8871) & 32.1359 (0.9031) & \textbf{32.3044} (\textbf{0.9055}) &  $\phantom{0}$58.99 & 31.9954 (0.9012)   & \textbf{32.2949} (\textbf{0.9057})&  $\phantom{0}$57.86\\
			& $\times4$ & 26.6742 (0.7278)  &  27.1836 (0.7434)  & \textbf{27.2993} (\textbf{0.7488}) & $\phantom{0}$42.87 & 27.1254 (0.7395) &\textbf{27.3040} (\textbf{0.7480}) &  $\phantom{0}$41.79\\
			\midrule
			\multirow{2}{*}{BSD100} & $\times2$ & 30.1373 (0.8671) & 31.1124 (0.8840)  & \textbf{31.2097} (\textbf{0.8864})&  $\phantom{0}$26.40 &  31.1087 (0.8835) & \textbf{31.2241} (\textbf{0.8866}) &   $\phantom{0}$27.13 \\
			& $\times4$ & 26.3402 (0.6900) &26.7099  (0.7027)  & \textbf{26.7891} (\textbf{0.7086}) &  $\phantom{0}$16.72 & 26.7027 (0.7018)&\textbf{26.7838} (\textbf{0.7085}) &  $\phantom{0}$16.90\\
			\midrule
			\multirow{2}{*}{Urban100}  & $\times2$ & 27.5143 (0.8728) & 28.7415 (0.8940)  & \textbf{28.8901} (\textbf{0.8953}) &  $\phantom{0}$37.16 & 28.6505 (0.8909) & \textbf{28.8415} (\textbf{0.8939}) &  $\phantom{0}$30.56  \\
			& $\times4$ & 23.7529 (0.6977) & 24.1968 (0.7104) & \textbf{24.2778} (\textbf{0.7147}) & 215.76 & 24.1443 (0.7047) & \textbf{24.2368} (\textbf{0.7114}) & 210.74 \\
			[\tbsp]
			\midrule[1.1pt]
			\textcolor{thisblue}{\sf \textbf{Dataset}} & 
			\textcolor{thisblue}{\sf \textbf{Scale}} & 
			\textcolor{thisblue}{\sf \textbf{TVAL3}} \cite{li_efficient_2013}  &  
			\textcolor{darkgreen}{\sf SelfExSR}~\cite{huang_single_2015} &
			\textcolor{darkredc}{\sf \textit{\textbf{Ours}}}                              &
			\textcolor{thisblue}{\sf \textbf{Time}} &
			\textcolor{darkgreen}{\sf FSRCNN}~\cite{dong_accelerating_2016} &
			\textcolor{darkredc}{\sf \textit{\textbf{Ours}}}  &
			\textcolor{thisblue}{\sf \textbf{Time}}
			\\
			\midrule
			\multirow{2}{*}{Set5} & $\times2$ & 34.0315 (0.9354) & 36.5001   (0.9537)  & \textbf{36.5321} (\textbf{0.9542}) &  $\phantom{0}$61.90 & 36.9912 (0.9556) & \textbf{37.0394} (\textbf{0.9559})&  $\phantom{0}$68.97 \\
			& $\times4$ &  29.1708 (0.8349) & 30.3317 (0.8623)  & \textbf{30.3370} (\textbf{0.8625}) & $\phantom{0}$44.66 & 30.7122 (0.8658) &  \textbf{30.8005} (\textbf{0.8691}) &  $\phantom{0}$50.29 \\
			\midrule
			\multirow{2}{*}{Set14} & $\times2$ & 31.0033 (0.8871) &  32.2272 (0.9036)   & \textbf{32.3951} (\textbf{0.9059}) &  $\phantom{0}$57.24 & 32.6515 (0.9089) &\textbf{32.6935 }(\textbf{0.9092}) &  $\phantom{0}$67.49 \\
			& $\times4$ & 26.6742 (0.7278) & 27.4014  (0.7518)   &\textbf{27.4730} (\textbf{0.7536}) &  $\phantom{0}$39.79 & 27.6179 (0.7550) & \textbf{27.6890} (\textbf{0.7574}) &  $\phantom{0}$52.47 \\
			\midrule
			\multirow{2}{*}{BSD100} & $\times2$ & 30.1373 (0.8671) &  31.1833 (0.8855)   & \textbf{31.2056} (\textbf{0.8862}) &  $\phantom{0}$27.00& 31.5075 (0.8905) & \textbf{31.5250} (\textbf{0.8907}) &  $\phantom{0}$26.50  \\
			& $\times4$ & 26.3402 (0.6900) & 26.8459 (\textbf{0.7108})&\textbf{26.8482} (\textbf{0.7108}) &  $\phantom{0}$16.25  &  26.9675 (0.7130)  & \textbf{27.0011} (\textbf{0.7149})& $\phantom{0}$20.76\\
			\midrule
			\multirow{2}{*}{Urban100}  & $\times2$ & 27.5143 (0.8728) & 29.3785 (0.9032)& \textbf{29.4578} (\textbf{0.9035}) & 298.04 & 29.8734 (0.9010) & \textbf{29.8926} (\textbf{0.9013}) & 263.39\\
			& $\times4$ & 23.7529 (0.6977) & 24.8241 (\textbf{0.7386}) &\textbf{24.8315} (0.7385) & 203.47 & 24.6196 (0.7270) & \textbf{24.6619} (\textbf{0.7297}) & 203.82\\
			[\tbsp]
			
			\midrule[1.1pt]
			\textcolor{thisblue}{\sf \textbf{Dataset}} & 
			\textcolor{thisblue}{\sf \textbf{Scale}} & 
			\textcolor{thisblue}{\sf \textbf{TVAL3}} \cite{li_efficient_2013} &  
			\textcolor{darkgreen}{\sf DRCN}~\cite{kim_DRCN} &
			\textcolor{darkredc}{\sf \textit{\textbf{Ours}}}                      &
			\textcolor{thisblue}{\sf \textbf{Time}}  	&
			\textcolor{darkgreen}{\sf VDSR}~\cite{Kim_VDSR} &
			\textcolor{darkredc}{\sf \textit{\textbf{Ours}}}                      &
			\textcolor{thisblue}{\sf \textbf{Time}}
			\\
			\midrule
			\multirow{2}{*}{Set5}& $\times2$ & 34.0315 (0.9354) & 37.6279 (0.9588) & \textbf{37.6712} (\textbf{0.9591})&  $\phantom{0}$57.23 & 37.5295 (0.9587)  &\textbf{37.5669} (\textbf{0.9588}) &$\phantom{0}$65.91 \\
			& $\times4$ & 29.1708 (0.8349) & 31.5344 (0.8854) &   \textbf{31.5701} (\textbf{0.8858}) &$\phantom{0}$46.66 & 31.3485 (0.8838) & \textbf{31.3780} (\textbf{0.8840}) & $\phantom{0}$47.59 \\
			\midrule
			\multirow{2}{*}{Set14} & $\times2$ & 31.0033 (0.8871)    &  33.0585 (0.9121) &\textbf{33.1038} (\textbf{0.9129)} 
			& $\phantom{0}$55.65 & 33.0527 (0.9127) & \textbf{33.1058} (\textbf{0.9131}) & $\phantom{0}$62.57\\
			& $\times4$ &  26.6742 (0.7278)  &  28.0269 (0.7673)  &\textbf{28.0588} (\textbf{0.7680}) & $\phantom{0}$42.57 &  28.0152 (0.7678)  & \textbf{28.0487} (\textbf{0.7684}) & $\phantom{0}$53.32\\
			\midrule
			\multirow{2}{*}{BSD100} & $\times2$ & 30.1373 (0.8671)& 31.8536 (0.8942) & \textbf{31.8737} (\textbf{0.8953}) &  $\phantom{0}$26.46 & 31.9015 (0.8960)  & \textbf{31.9142} (\textbf{0.8962}) & $\phantom{0}$26.92 \\
			& $\times4$ & 26.3402 (0.6900) & 27.2364 (0.7233) & \textbf{27.2524} (\textbf{0.7240}) & $\phantom{0}$19.17  & 26.8776 (0.7093) &  \textbf{27.2535} (\textbf{0.7238}) & $\phantom{0}$20.74 \\
			[\tbsp]
			\midrule[1.1pt]
			\textcolor{thisblue}{\sf \textbf{Dataset}} & 
			\textcolor{thisblue}{\sf \textbf{Scale}}  & 
			\textcolor{thisblue}{\sf \textbf{TVAL3}} \cite{li_efficient_2013} &
			\textcolor{darkgreen}{\sf LapSRN}~\cite{lai_deep_2017} &
			\textcolor{darkredc}{\sf \textit{\textbf{Ours}}}   &
			\textcolor{thisblue}{\sf \textbf{Time}} &
			\textcolor{darkgreen}{\sf SRMD}~\cite{zhang_learning_2018} &
			\textcolor{darkredc}{\sf \textit{\textbf{Ours}}} &
			\textcolor{thisblue}{\sf \textbf{Time}}
			\\
			\midrule
			\multirow{3}{*} {Set5} & $\times2$ & 34.0315 (0.9354) & 37.7008 (0.9590) & \textbf{37.7219} (\textbf{0.9592}) & $\phantom{0}$62.45 &  37.4496 (0.9579) & \textbf{37.5817} (\textbf{0.9585}) &$\phantom{0}$74.33 \\
			& $\times4$ & 29.1708 (0.8349) &  31.7181 (0.8891) & \textbf{31.7428} (\textbf{0.8894}) & $\phantom{0}$45.19 &  31.5750 (0.8853) & \textbf{31.6531} (\textbf{0.8863}) & $\phantom{0}$58.01\\
			& $\times 8$ & 24.9663 (0.6722) & 26.3314 (\textbf{0.7548}) & \textbf{26.3881} (0.7545) & $\phantom{0}$50.01\\
			\midrule
			\multirow{3}{*}{Set14} & $\times2$ & 31.0033 (0.8871)  & 33.2518 (0.9138) & \textbf{33.2709} (\textbf{0.9142}) & $\phantom{0}$56.65 & 33.1035 (0.9127) &\textbf{33.1868} (\textbf{0.9137})& $\phantom{0}$69.58\\
			& $\times4$ & 26.6742 (0.7278) & 28.2533 (0.7730)  & \textbf{28.2722} (\textbf{0.7734}) &$\phantom{0}$42.04 & 28.1593 (0.7716) & \textbf{28.2174} (\textbf{0.7728}) & $\phantom{0}$52.03\\
			& $\times 8$ &23.6079 (0.5800) &24.5643 (\textbf{0.6266}) & \textbf{24.5993} (0.6264) &$\phantom{0}$42.48\\
			\midrule
			\multirow{3}{*}{BSD100} & $\times2$ & 30.1373 (0.8671) & 32.0214 (0.8970)  & \textbf{32.0274} (\textbf{0.8975}) & $\phantom{0}$24.53 & 31.8722 (0.8953)  &  \textbf{31.9009} (\textbf{0.8959}) &$\phantom{0}$28.15 \\
			& $\times4$ & 26.3402 (0.6900) & 27.4164 (0.7296) & \textbf{27.4317} (\textbf{0.7300}) & $\phantom{0}$19.03 & 27.3350 (0.7273) & \textbf{27.3579} (\textbf{0.7280})  & $\phantom{0}$20.94\\
			& $\times 8$ & 23.9971 (0.5542) &  24.6495 (\textbf{0.5887})  & \textbf{24.6769} (0.5886) &$\phantom{0}$18.30\\
			\midrule
			\multirow{3}{*}{Urban100}  & $\times2$ &  27.9935 (0.8742) &  31.1319 (0.9180) & \textbf{31.1462} (\textbf{0.9183}) &  259.22 &  30.8799 (0.9146)&  \textbf{30.9253} (\textbf{0.9151}) &  266.98 \\
			& $\times4$ &  23.7529 (0.6977) & 25.5026 (0.7661) & \textbf{25.5167} (\textbf{0.7662}) & 199.70 & 25.3494 (0.7605) & \textbf{25.3834} (\textbf{0.7609}) & 209.77 \\ 
			& $\times 8$ & 21.1208 (0.5357) & 22.0547 (\textbf{0.5956}) & \textbf{22.0675} (0.5944) & 178.60 \\[\tbsp]
			
			\midrule[1.1pt]
			\textcolor{thisblue}{\sf \textbf{Dataset}}                     & 
			\textcolor{thisblue}{\sf \textbf{Scale}}                       & 
			\textcolor{thisblue}{\sf \textbf{TVAL3}} \cite{li_efficient_2013}          &
			\textcolor{darkgreen}{\sf IRCNN}~\cite{IRCNN-zhang2017} &
			\textcolor{darkredc}{\sf \textit{\textbf{Ours}}}   &
			\textcolor{thisblue}{\sf \textbf{Time}} &
			\textcolor{darkgreen}{\sf ESRGAN}~\cite{wang2018esrgan} &
			\textcolor{darkredc}{\sf \textit{\textbf{Ours}}}                         &
			\textcolor{thisblue}{\sf \textbf{Time}} 
			\\
			\midrule
			\multirow{2}{*} {Set5} & 	$\times2$ & 34.0315 (0.9354) & 37.3436 (0.9572) & \textbf{37.3684} (\textbf{0.9576}) & $\phantom{0}$67.03 &  &   &  \\ 
			& $\times4$ &  29.1708 (0.8349) & 30.9995 (0.8778)& \textbf{31.0041} (\textbf{0.8779})& $\phantom{0}$52.06& 32.7072 (0.9001) &  \textbf{32.7170} (\textbf{0.9002}) & $\phantom{0}$44.33\\
			\midrule
			\multirow{2}{*}{Set14} & $\times2$ & 31.0033 (0.8871) & 32.8573 (0.9105) & \textbf{32.8929} (\textbf{0.9110}) & $\phantom{0}$65.13&  &   &\\
			& $\times4$ & 26.6742 (0.7278)  & 27.7195 (0.7614)& \textbf{27.7420} (\textbf{0.7615})& $\phantom{0}$52.36& 28.8342 (0.7877)  & \textbf{28.9253} (\textbf{0.7891}) &$\phantom{0}$42.40 \\
			\midrule
			\multirow{2}{*}{BSD100}& $\times2$ & 31.0033 (0.8871) & 31.6543 (0.8918) &  \textbf{31.6723} (\textbf{0.8923}) & $\phantom{0}$25.04 &  &  & \\
			& $\times4$ & 26.3402 (0.6900) & 27.0848 (0.7188)& \textbf{27.0922} (\textbf{0.7189}) & $\phantom{0}$20.79 &  27.8332 (\textbf{0.7447})  & \textbf{27.8489} (\textbf{0.7447}) &  $\phantom{0}$19.05\\
			\midrule
			\multirow{2}{*}{Urban100} & $\times2$ & 31.0033 (0.8871) &  30.0623 (0.9105)&  \textbf{30.0806} (\textbf{0.9108}) & $\phantom{0}$74.41&  &   & \\
			& $\times4$ & 23.7529 (0.6977) & 24.8913 (0.7395) & \textbf{24.9007} (\textbf{0.7396}) &  212.49 & 27.0270 (\textbf{0.8146}) & \textbf{27.0404} (\textbf{0.8146}) & 206.61\\
			\midrule[1.1pt]
		\end{tabular}
	}
  \label{tab:3}
    \end{table*}
}
\isdraft{
\label{SubFig:baboonGT}
\begin{figure*}[!t]
\includegraphics[scale = 0.9]{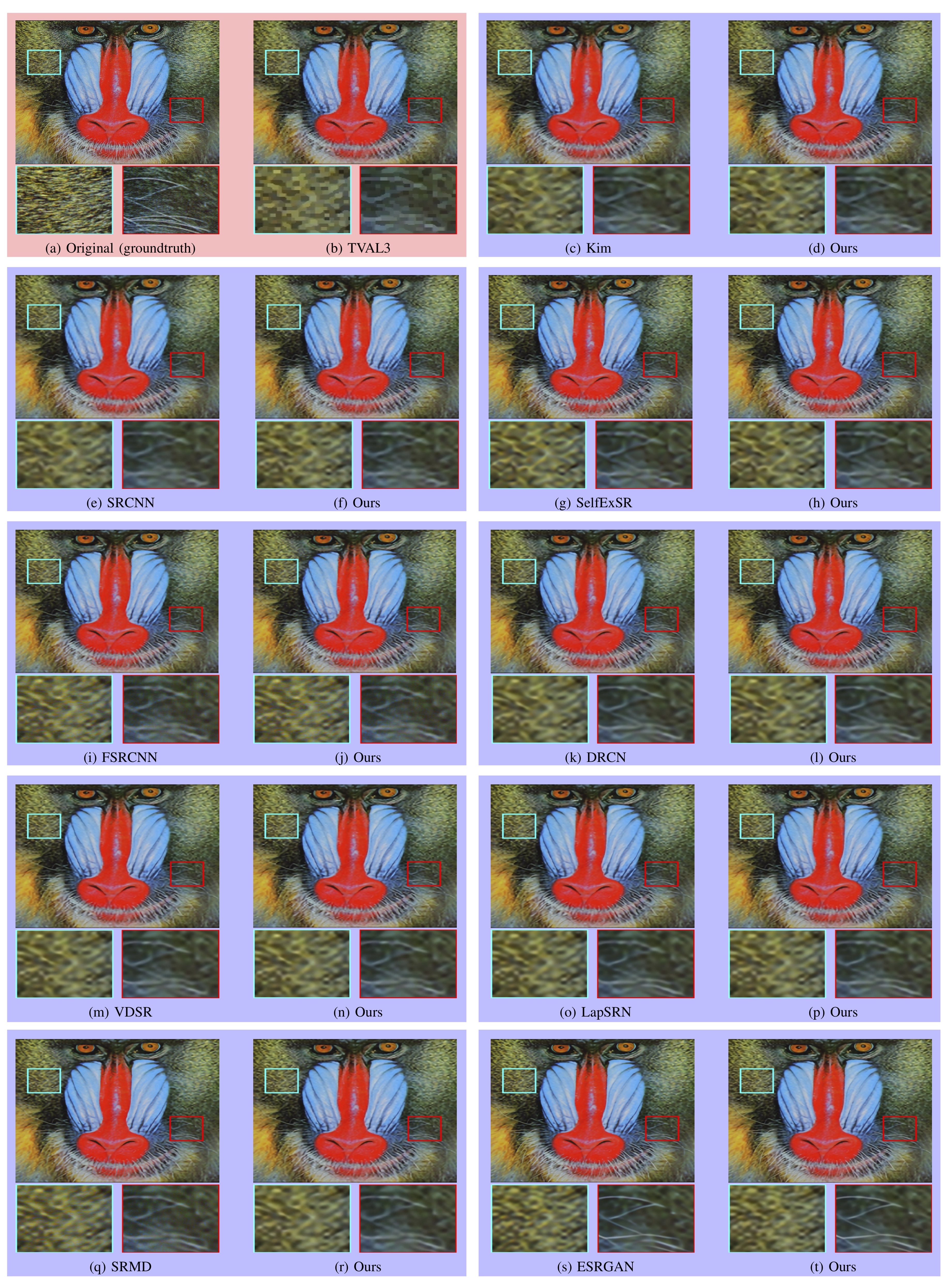}
\caption{
 \footnotesize
   Results on \textit{Baboon} (Set14) for $4\times$. Each shaded area (except
  the top-left) shows the output of a learning-based algorithm and of our method.}
  \label{fig_results1}
 \end{figure*}
}{
\begin{figure*}[t]
	\centering
\includegraphics[scale = 0.13]{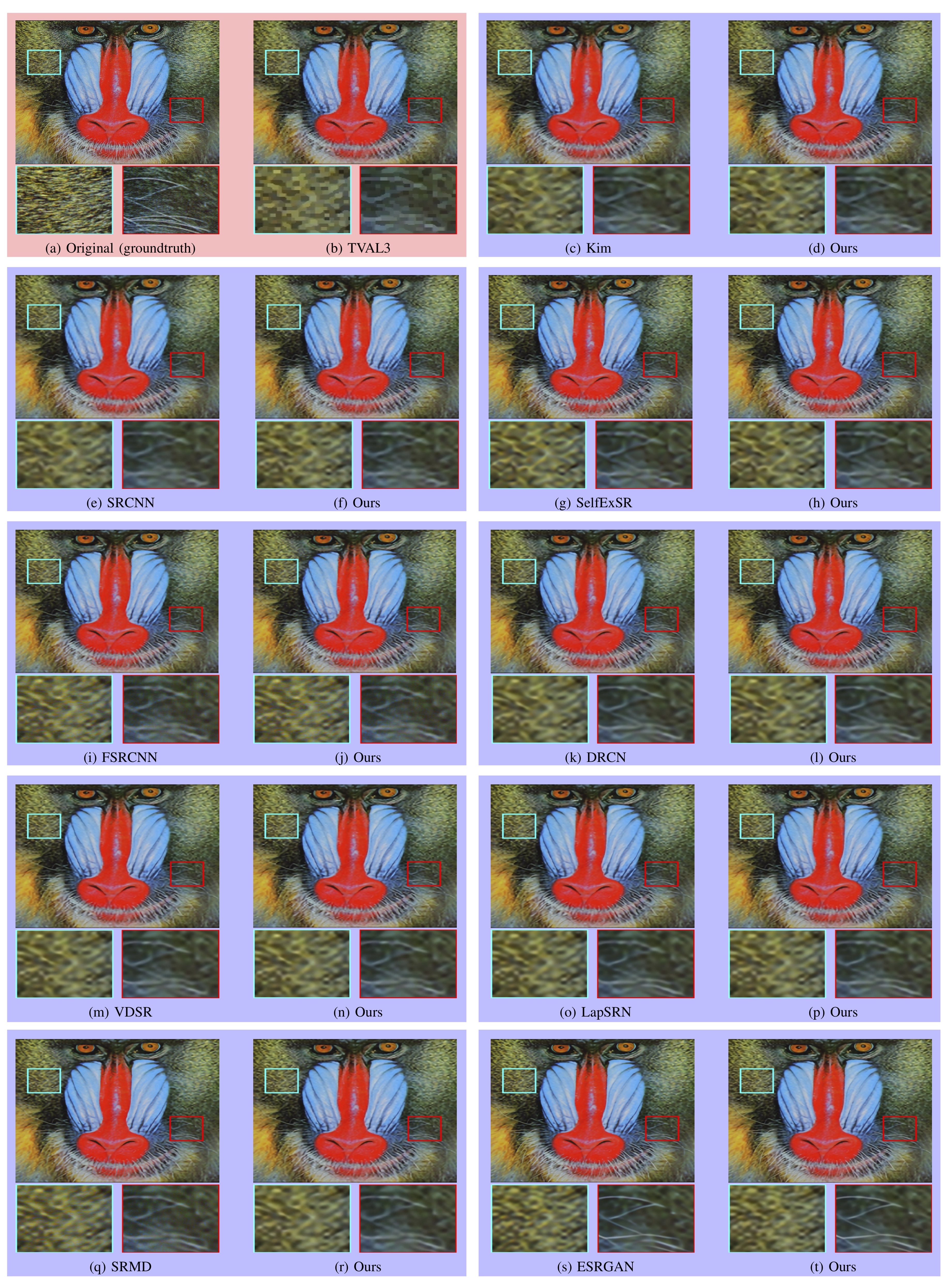}
\caption{
	\footnotesize
	Results on \textit{Baboon} (Set14) for $4\times$. Each shaded area (except
	the top-left) shows the output of a learning-based algorithm and of our method.}
\label{fig_results1}
\end{figure*} 
}
\isdraft{
\begin{figure*}[!t]
\includegraphics[height = 20cm, width = 18cm]{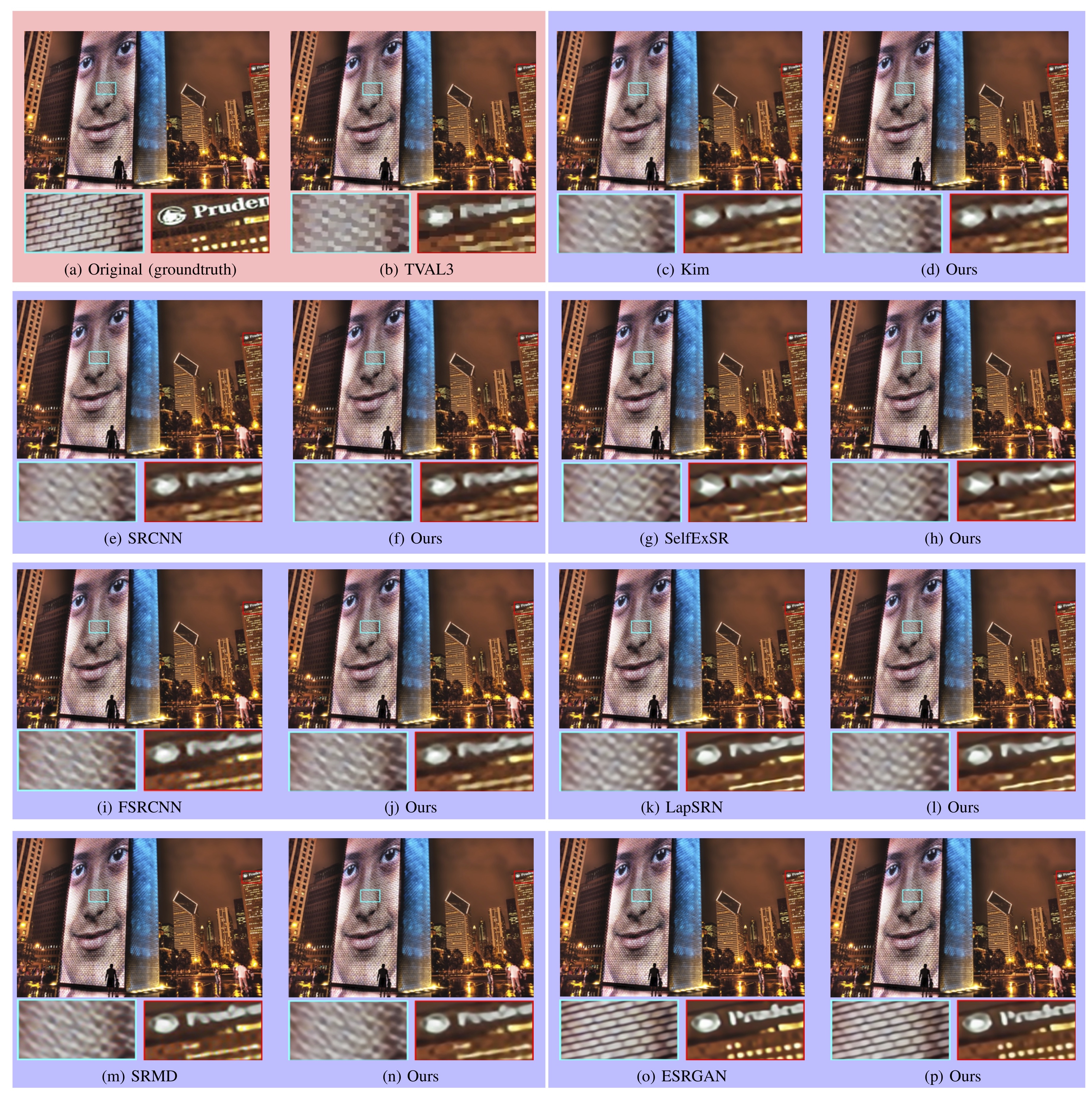}
\caption{
 \footnotesize
  Results on \textit{img076} image from Urban100 for $4\times$. Each shaded area (except the top-left) shows the output of a learning-based algorithm and of our method.}
 \label{fig_results2}
 \end{figure*}
}{
\begin{figure*}[t]
\includegraphics[height = 20cm, width = 18cm]{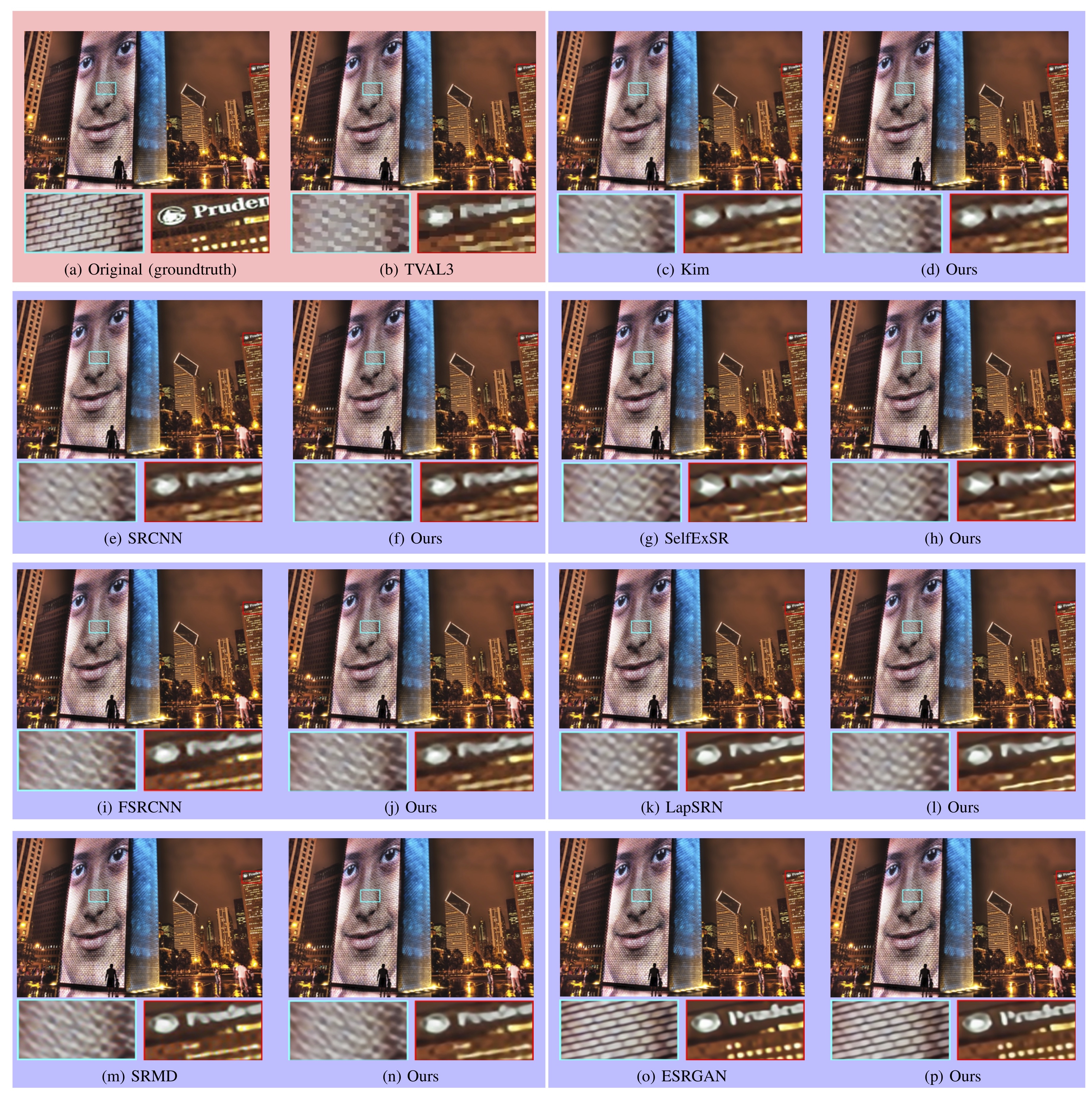}
\caption{
	\footnotesize
	Results on \textit{img076} image from Urban100 for $4\times$. Each shaded area (except the top-left) shows the output of a learning-based algorithm and of our method.}
\label{fig_results2}
\end{figure*}
}
\subsection{Standard Datasets With Bicubic Downsampling}

We also conducted more systematic experiments using the standard datasets Set5,
Set14, BSD100, and Urban100, under different scaling factors and using
bicubic downsampling only.

\mypar{Quantitative results}
Table~\ref{tab:3} displays the average PSNR and SSIM, as well as the average
execution time of our method (in seconds), for $2\times$, $4\times$, and
$8\times$ scaling factors. Each shaded area shows the performance of a given
(learning-based) reference method, the performance of our scheme applied to the
output of that reference method, and the average execution time (of our
method). For easy comparison, the values for TVAL3 occur repeatedly in different vertical sub-blocks
of the table. Note that the 4th vertical sub-block of the table
is the only one with values for $8\times$ upsampling, since LapSRN is the only
method that can handle such a scaling
factor. Also, as ESRGAN was designed specifically for $4\times$ upsampling, we
do not present its values for other scaling factors. All results for our method
were generated with $\beta = 1$ in~\eqref{eq:TV-TV}, except when we considered
LapSRN and ESRGAN for the Urban100 dataset, in which case we set $\beta = 2$.

An obvious pattern in the table is that our method consistently improves the
outputs of all the methods in terms of PSNR and SSIM, except in a small subset
of cases. The improvements range between 0.0023 and 0.3796 dB. One of the
exceptions occurs for $8\times$ upsampling with LapSRN. In that case, LapSRN has always better SSIM
than our method, even though the opposite happens for the PSNR. As expected,
TVAL3 had the worst performance  overall and was surpassed by all learning-based methods. 

A drawback of our method, however, is its possibly long execution time.  We
recall that of all downsampling operators mentioned in
Section~\ref{SubSec:ModelAndAssumptions}, the most computationally complex is
bicubic downsampling, as considered in these experiments. The timing values in
Table~\ref{tab:3} are average values: they report the total execution time of
our algorithm over all the images of the corresponding dataset divided by the
number of images. While in some cases our algorithm took an average of 16 sec
(SRCNN, BSD100, $4\times$), in others it took more than 250 sec (LapSRN,
Urban100, $2\times$). In fact, for the Urban100 dataset, because of the large
size of its images ($1024 \times 644$), we had to reduce the number of
simultaneous threads to prevent the GPUs from overflowing. For reference, for (SRCNN, BSD100, $4\times$), our method takes an average of 9 sec when we use simple subsampling. This is roughly half the execution time it takes for bicubic downsampling.  

\mypar{Qualitative results}
Figures~\ref{fig_results1} and~\ref{fig_results2} depict the output images of
all the algorithms (except IRCNN) for the test images \textit{baboon} from
Set14, and \textit{img067} from Urban100. All super-resolved images exhibit
blur and loss of information compared with the groundtruth images in
Figs.\isdraft{~3a-4a}{~\ref{SubFig:baboonGT}-\ref{SubFig:076GT}}. 
And as our scheme builds upon the outputs of other methods, it also inherits
some of their artifacts. It is difficult to visually assess differences between
the outputs of the algorithms and of our method, in part because the
improvements, as measured by the PSNR, are relatively small. Yet, as our
experiments show, our scheme not only systematically improves the outputs of
CNN-based methods, but also adds significant robustness to operator mismatch. 

\section{Conclusions}
\label{Section5}

We proposed a framework for single-image super-resolution that blends model- and learning-based (e.g., CNN) techniques. As a result, our framework enables solving the consistency problem that CNNs suffer from, namely that downsampled output (HR) images fail to match the input (LR) images. Our experiments show that enforcing such consistency not only systematically improves the quality of the output images of CNNs, but also adds robustness to the super-resolution task. At the core of our framework is a problem that we call TV-TV minimization and which we solve with an ADMM-based algorithm. Although our implementation is efficient, there is still margin to improve its execution time, for example, by unrolling the proposed algorithm with a neural network. We leave this for future work.

\section*{Acknowledgments}

Work supported in part by UK's EPSRC (EP/S000631/1), the UK MOD University Defence
Research Collaboration. Computational resources provided by EPSRC Capital
Award (EP/S018018/1).


\bibliographystyle{IEEEtran}
\bibliography{RefList}

\end{document}